\documentclass{article}
\usepackage{arxiv}
\usepackage[utf8]{inputenc} % allow utf-8 input
\usepackage[T1]{fontenc}    % use 8-bit T1 fonts
\usepackage{hyperref}       % hyperlinks
\usepackage{url}            % simple URL typesetting
\usepackage{booktabs}       % professional-quality tables
\usepackage{amsfonts}       % blackboard math symbols
\usepackage{nicefrac}       % compact symbols for 1/2, etc.
\usepackage{microtype}      % microtypography
\usepackage{graphicx}
\graphicspath{ {./images/} }
\usepackage{longtable}
\usepackage{caption}
\usepackage{geometry}
\usepackage{fancyhdr}
\usepackage{afterpage}
\usepackage{subfigure}
\usepackage{amsmath,amssymb,amsbsy}
\usepackage{algorithmicx}
\usepackage{algorithm}
\usepackage{algpseudocode}
\usepackage{dcolumn,array}
\usepackage{subcaption}
\usepackage{lscape}
\usepackage{multirow}

\title{A Generalized Framework for Multiscale State-Space Modeling with Nested Nonlinear Dynamics: An Application to Bayesian Learning under Switching Regimes}

\author{
 Nayely V\'{e}lez-Cruz \\
  School of Complex Adaptive Systems\\
Arizona State University\\
  Tempe, AZ 85282 \\
  \texttt{nvelezcr@asu.edu} \\
  \And
 Manfred D. Laubichler \\
  School of Complex Adaptive Systems\\
  Arizona State University\\
  Tempe, AZ 85282 \\
  \texttt{Manfred.Laubichler@asu.edu} \\
}

\begin{document}
\maketitle

\begin{abstract}
In this work, we introduce a generalized framework for multiscale state-space models that incorporates nested nonlinear dynamics, with a specific focus on Bayesian learning under switching regimes. Our framework captures the complex interactions between fast and slow processes within systems, allowing for the analysis of how these dynamics influence each other across various temporal scales. We model these interactions through a hierarchical structure in which finer time-scale dynamics are nested within coarser ones, facilitating feedback between the scales. To promote the practical application of our framework, we address the problem of identifying switching regimes and transient dynamics. In particular, we develop a Bayesian learning approach to estimate latent states and indicators corresponding to switching dynamics, enabling the model to adapt effectively to regime changes. We employ Sequential Monte Carlo, or particle filtering, for inference. We illustrate the utility of our framework through simulations. The results demonstrate that our Bayesian learning approach effectively tracks state transitions and achieves accurate identification of switching dynamics in multiscale systems.
\end{abstract}

\section{Introduction}
In complex systems, processes operate across multiple time scales, such as rapid fluctuations in environmental conditions, intermediate responses like population dynamics, and slower shifts such as ecosystem succession or climate change. These dynamics are often nested, with fast processes embedded within slower ones. Fine-scale, rapid changes can accumulate over time to influence large-scale trends, while slower processes provide the conditions for fast dynamics to unfold.

This interplay between processes at different time scales can lead to transient behaviors, where a system remains in one dynamic state for an extended period before abruptly shifting to another \cite{hastings2018transient}. In ecological systems, these dynamics often manifest as long transients—periods of apparent stability followed by sudden regime shifts. These shifts can occur without any obvious external trigger, driven instead by internal processes or responses to environmental variability \cite{scheffer2001catastrophic}. During these phases, a system may exhibit consistent behavior over time before transitioning to a different dynamic regime, which could involve altered oscillatory patterns or a completely new structure. Such transitions are difficult to predict, as they are nonlinear, involve systems operating at multiple interacting scales, and are influenced by stochasticity \cite{morozov2020transient, dakos2012methods}.

Understanding these multiscale and nonlinear interactions is essential for anticipating regime shifts, which are often most consequential at the coarsest time scales, where changes in slow-moving processes like ecosystem succession or long-term climate changes lead to impactful, irreversible transitions \cite{scheffer2001catastrophic}. In contrast, finer time scales capture rapid, fluctuations, such as daily population variations or nutrient cycles, which  do not induce permanent changes unless their effects accumulate over time.

Recognizing the critical roles these dynamics play informs our approach to modeling them effectively. In this work, we propose a generalized multiscale state-space modeling framework with nested nonlinear dynamics. This framework, which is an extension of our work in \cite{velez2024bayesian}, is designed to incorporate interactions between any number of temporal scales, enabling a comprehensive representation of complex systems where localized, rapid changes can propagate and influence slower, broader-scale phenomena and vice versa. In addition to modeling interactions across time scales, we also account for switching dynamics within the broadest time scale. In particular, we introduce a Bayesian learning approach to estimate the latent states and identify the underlying dynamic regimes, allowing for flexible modeling of systems where the dynamics may switch between different models over time.

\section{Model Description}
At each time scale \(l\), the latent state $\mathbf{x}^{[l]}_{d,t_l}$ for individual \(d\) evolves based on its previous state $\mathbf{x}^{[l]}_{d,t_l-1}$, the entire trajectory of the finer time scales $\left\{ \mathbf{x}^{[l']}_{d,1:T_{l'}} \right\}_{l'=1}^{l-1}$, and the states from the coarser time scales $\left\{ \mathbf{x}^{[l']}_{d,t_{l'}-1} \right\}_{l'=l+1}^{L}$. The dependence on the coarser time scale states $\mathbf{x}^{[l']}_{d,t_{l'}-1}$ reflects the influence of slower, large-scale processes, which can shape the overall context in which finer-scale dynamics unfold, providing broader temporal trends or constraints from longer time horizons. This structure allows feedback across time scales, enabling interactions between processes operating at different temporal resolutions. Additionally, the model includes additive process noise, $\mathbf{w}^{[l]}_{d,t_{l}-1}$, to account for stochastic influences. 
The state transition equation for time scale \(l\) and individual (or more generally, entity) \(d\), is given by: 
\begin{align}
\mathbf{x}^{[l]}_{d,t_l} &= f_l\left( \mathbf{x}^{[l]}_{d,t_l-1}, \left\{ \mathbf{x}^{[l']}_{d,1:T_{l'}} \right\}_{l'=1}^{l-1}, \left\{ \mathbf{x}^{[l']}_{d,t_{l'}-1} \right\}_{l'=l+1}^{L} \right) +  \mathbf{w}^{[l]}_{d,t_l-1} \\
\mathbf{y}^{[l]}_{d,t_l} &= h_l(\mathbf{x}^{[l]}_{d,t_l}) + \mathbf{v}^{[l]}_{d,t_l}
\end{align}
The function \(f_l\) captures the transition dynamics for the latent state at time scale \(l\), incorporating the previous state, the entire trajectory from all finer time scales, and additive process noise $\mathbf{w}^{[l]}_{d,t_{l}-1}$. The measurement model equation for \(\mathbf{y}^{[l]}_{d,t_l}\) describes how the measurements at time scale \(l\) (denoted as \(\mathbf{y}^{[l]}_{d,t_l}\)) relates to the latent state \(\mathbf{x}^{[l]}_{d,t_l}\). The function \(h_l(\mathbf{x}^{[l]}_{d,t_l})\) maps the latent state to the measurements, while \(\mathbf{v}^{[l]}_{d,t_l}\) represents the measurement noise that accounts for measurement errors or uncertainties in the measurements.

\begin{longtable}{|>{\centering\arraybackslash}m{3cm}|>{\centering\arraybackslash}m{9cm}|}
\caption{Notation Table} \label{tab:notation}\\
\hline
\textbf{Variable} & \textbf{Description} \\
\hline
\(\mathbf{x}^{[l]}_{d,t_l}\) & Latent state of individual \(d\) at time scale \(l\) and time \(t_l\). \\
\hline
\(\left\{ \mathbf{x}^{[l']}_{d,1:T_{l'}} \right\}_{l'=1}^{l-1}\) & The entire trajectory of states from all finer time scales, representing the influences across time scales. \\
\hline
\(\left\{ \mathbf{x}^{[l']}_{d,t_{l'}-1} \right\}_{l'=l+1}^{L}\) & The set of states from all coarser time scales at the previous coarse time point, capturing the broader, slower dynamics that influence the finer time scale processes. \\
\hline
\(f_l(\cdot)\) & Transition function for the latent state at time scale \(l\), incorporating feedback from finer and coarser time scales. \\
\hline
\(B^{[l]}_{d,d'}\) & Interaction matrix capturing the interactions between individuals \(d\) and \(d'\) at time scale \(l\). \\
\hline
\(\mathbf{A}^{[l]}_{t_l}\) & Adjacency matrix capturing interactions between dimensions of the state vector for individual \(d\) at time \(t_l\). \\
\hline
\(\mathbf{w}^{[l]}_{d,t_{l}-1}\) & Process noise for state transition at time scale \(l\), representing random fluctuations in the system at time \(t_l\). \\
\hline
\(\mathbf{y}^{[l]}_{d,t_l}\) & Measurements at time scale \(l\), measured at time \(t_l\). \\
\hline
\(h_l(\cdot)\) & Measurement function mapping the latent state at time scale \(l\) to the measurements. \\
\hline
\(\mathbf{v}^{[l]}_{d,t_l}\) & Measurement noise at time scale \(l\), accounting for measurement errors or noise in the measurements at time \(t_l\). \\
\hline
\(t_l\) & Time index corresponding to the time scale \(l\), representing different resolutions of time depending on \(l\). \\
\hline
\(L\) & The total number of time scales in the model, with the finest time scale \(l = 1\) and the coarsest \(l = L\). \\
\hline
\(l'\) & An index representing all time scales from \(1\) to \(l-1\), referring to finer time scales. \\
\hline
\end{longtable}

\subsection{Example: \(L = 2\)}
To illustrate the model, consider the case of \(L = 2\), where there are two time scales. The finest time scale \(l = 1\) may represent fast processes such as development, while the coarser time scale \(l = 2\) may represent slower processes such as heredity or population dynamics. At the finest time scale \(l = 1\), the latent state evolves as:
\begin{equation}
\mathbf{x}^{[1]}_{d,t_1} = f_1\left( \mathbf{x}^{[1]}_{d,t_1-1}, \mathbf{x}^{[2]}_{d,t_{2}-1} \right) + \mathbf{w}^{[1]}_{d,t_{1}-1}
\end{equation}
At the coarser time scale \(l = 2\), the latent state depends on the full trajectory of the finer time scale \(l = 1\), evolving as:
\begin{equation}
\mathbf{x}^{[2]}_{d,t_2} = f_2\left( \mathbf{x}^{[2]}_{d,t_2-1}, \mathbf{x}^{[1]}_{d,1:T_1} \right) + \mathbf{w}^{[2]}_{d,t_{2}-1}
\end{equation}
The measurement at each time scale follows:
\begin{align}
\mathbf{y}^{[1]}_{d,t_1} &= h_1\left( \mathbf{x}^{[1]}_{d,t_1} \right) + \mathbf{v}^{[1]}_{d,t_1} \\
\mathbf{y}^{[2]}_{d,t_2} &= h_2\left( \mathbf{x}^{[2]}_{d,t_2} \right) + \mathbf{v}^{[2]}_{d,t_2}
\end{align}

\subsection{Example: \(L = 3\)}
To illustrate the model with three time scales, consider the case of \(L = 3\), where there are three nested temporal scales. The finest time scale \(l = 1\) may represent fast processes such as cellular signaling, the intermediate time scale \(l = 2\) may represent developmental processes, and the coarsest time scale \(l = 3\) may represent evolutionary or population-level dynamics. At the finest time scale \(l = 1\), the latent state evolves as:
\begin{equation}
\mathbf{x}^{[1]}_{d,t_1} = f_1\left( \mathbf{x}^{[1]}_{d,t_1-1}, \mathbf{x}^{[2]}_{d,t_{2}-1}, \mathbf{x}^{[3]}_{d,t_{3}-1} \right) + \mathbf{w}^{[1]}_{d,t_{1}-1}
\end{equation}
At the intermediate time scale \(l = 2\), the latent state depends on the full trajectory of the finer time scale \(l = 1\), evolving as:
\begin{equation}
\mathbf{x}^{[2]}_{d,t_2} = f_2\left( \mathbf{x}^{[2]}_{d,t_2-1}, \mathbf{x}^{[1]}_{d,1:T_1}, \mathbf{x}^{[3]}_{d,t_{3}-1} \right) + \mathbf{w}^{[2]}_{d,t_{2}-1}
\end{equation}

At the coarsest time scale \(l = 3\), the latent state is influenced by both finer time scales, evolving as:
\begin{equation}
\mathbf{x}^{[3]}_{d,t_3} = f_3\left( \mathbf{x}^{[3]}_{d,t_3-1}, \mathbf{x}^{[1]}_{d,1:T_1}, \mathbf{x}^{[2]}_{d,1:T_2} \right) + \mathbf{w}^{[3]}_{d,t_{3}-1}
\end{equation}
The measurements at each time scale are modeled as:
\begin{align}
\mathbf{y}^{[1]}_{d,t_1} &= h_1\left( \mathbf{x}^{[1]}_{d,t_1} \right) + \mathbf{v}^{[1]}_{d,t_1} \\
\mathbf{y}^{[2]}_{d,t_2} &= h_2\left( \mathbf{x}^{[2]}_{d,t_2} \right) + \mathbf{v}^{[2]}_{d,t_2} \\
\mathbf{y}^{[3]}_{d,t_3} &= h_3\left( \mathbf{x}^{[3]}_{d,t_3} \right) + \mathbf{v}^{[3]}_{d,t_3}
\end{align}

\subsection{Introducing Interaction Matrices \(A\) and \(B\)}
To capture the dependencies between different dimensions of the state vector within the same individual at time \( t_l \), we introduce an adjacency matrix \( \mathbf{A}^{[l]}_{t_l} \). This matrix effectively models the interactions among the elements of the state vector \( \mathbf{x}^{[l]}_{d,t_l} \), allowing us to account for how each state component influences others at a given time. For instance, in a biological system, the state components might represent different gene expression levels, where the adjacency matrix reflects regulatory relationships among genes. For finer time scales \( l < L \), the state transition model is given by:
\begin{equation}
\mathbf{x}^{[l]}_{d,t_l} = f_l\left( \mathbf{A}^{[l]}_{t_l}\mathbf{x}^{[l]}_{d,t_l-1}, \left\{ \mathbf{x}^{[l']}_{d,1:T_{l'}} \right\}_{l'=1}^{l-1} , \left\{ \mathbf{x}^{[l']}_{d,t_{l'}-1} \right\}_{l'=l+1}^{L}\right) +  \mathbf{w}^{[l]}_{d,t_{l}-1}
\end{equation}
Next, to account for interactions between different individuals \( d \) and \( d' \) at the coarsest time scale \( L \), we introduce an interaction matrix \( \mathbf{B}^{[L]} \). This matrix influences the evolution of the latent states by capturing how the states of different individuals affect each other. For example, in social networks, the interaction matrix can represent the influence one individual's state may have on another. At the coarsest time scale \( l = L \), the updated state transition model incorporates the interaction matrix \( \mathbf{B}^{[L]} \):
\begin{equation}
\mathbf{x}^{[L]}_{d,t_L} = f_L\left( \mathbf{A}^{[L]}_{t_L}\mathbf{x}^{[L]}_{d,t_L-1}, \left\{ \mathbf{x}^{[l']}_{d,1:T_{l'}} \right\}_{l'=1}^{L-1}, \sum_{d' \neq d} B^{[L]}_{d,d'} \mathbf{x}^{[L]}_{d',t_L-1} \right) +  \mathbf{w}^{[L]}_{d,t_{L}-1}
\end{equation}

\section{Bayesian Learning Under Switching Regimes}
We propose a Bayesian approach for learning the unknown states at different time scales, while restricting the switching regime to the coarsest time scale \( L \). Specifically, let \( s_{d,t_{L}}^{[L]} = m \) for \( m=1,...,M \) denote the indicator corresponding to model (or regime) \( m \) for time scale \( L \) at time step \( t_{L} \) for individual \( d \). The objective is to estimate the unknown states \( \mathbf{x}^{[l]}_{d,t_{l}} \) for \( t_{l} = 1,...,T_{l} \) and \( l=1,...,L \), and the unknown model indicators \( s_{d,t_{L}}^{[L]} \) at the coarsest time scale \( L \). For the finer time scales \( l < L \), the state evolves as:
\begin{align}
\mathbf{x}^{[l]}_{d,t_l} = f_l\left( \mathbf{A}^{[l]}_{t_l}\mathbf{x}^{[l]}_{d,t_l-1}, \left\{ \mathbf{x}^{[l']}_{d,1:T_{l'}} \right\}_{l'=1}^{l-1}, \left\{ \mathbf{x}^{[l']}_{d,t_{l'}-1} \right\}_{l'=l+1}^{L} \right) +  \mathbf{w}^{[l]}_{d,t_{l}-1}
\end{align}

At the coarsest time scale \( l = L \), the state evolves according to the model corresponding to $s_{d,t_{L}}^{[L]} = m$ :
\begin{align}
\mathbf{x}^{[L]}_{d,t_L} = f_L\left( \mathbf{A}^{[L]}_{t_L}\mathbf{x}^{[L]}_{d,t_L-1}, \left\{ \mathbf{x}^{[l']}_{d,1:T_{l'}} \right\}_{l'=1}^{L-1}, s_{d,t_{L}}^{[L]} = m, \sum_{d' \neq d} B^{[L]}_{d,d'} \mathbf{x}^{[L]}_{d',t_L-1} \right) +  \mathbf{w}^{[L]}_{d,t_{L}-1}
\end{align}
The measurement equation is assumed to be 
\begin{equation}
\mathbf{y}^{[l]}_{d,t_l} = h_l(\mathbf{x}^{[l]}_{d,t_l}) + \mathbf{v}^{[l]}_{d,t_l}
\end{equation}
Note that for $l=1,...,L$ $\mathbf{w}^{[l]}_{d,t_{l}} \sim \mathcal{N}(\boldsymbol{0}, \Sigma^{[l]}_{\textbf{w}_{d}^{[l]}})$ and $\mathbf{v}^{[l]}_{d,t_{l}} \sim \mathcal{N}(\boldsymbol{0}, \Sigma^{[l]}_{\textbf{v}_{d}^{[l]}})$ are additive Gaussian noise terms. The function \( h_l \) applies a rotation matrix, transforming the state \( \mathbf{x}^{[l]}_{d,t_l} \) to the measurement space. 

This model is summarized by the following hierarchy:
\begin{align*}
    \textbf{x}_{d,t_{l}}^{[l]} &\vert \textbf{x}_{d,t_{l}-1}^{[l]}, \left\{ \mathbf{x}^{[l']}_{d,1:T_{l'}} \right\}_{l'=1}^{l-1},  \left\{ \mathbf{x}^{[l']}_{d,t_{l'}-1} \right\}_{l'=l+1}^{L} \sim \mathcal{N}\left( \textbf{x}_{d,t_{l}}^{[l]} \mid \textbf{x}_{d,t_{l}-1}^{[l]}, \left\{ \mathbf{x}^{[l']}_{d,1:T_{l'}} \right\}_{l'=1}^{l-1}, \left\{ \mathbf{x}^{[l']}_{d,t_{l'}-1} \right\}_{l'=l+1}^{L} \Sigma^{[l]}_{\textbf{w}_{d}^{[l]}} \right), \quad \text{for} \, l < L \\
    \textbf{x}_{d,t_{L}}^{[L]} &\vert \textbf{x}_{d,t_{L}-1}^{[L]}, \left\{ \mathbf{x}^{[l']}_{d,1:T_{l'}} \right\}_{l'=1}^{L-1}, s_{d,t_{L}}^{[L]} = m \sim \mathcal{N}\left( \textbf{x}_{d,t_{L}}^{[L]} \mid \textbf{x}_{d,t_{L}-1}^{[L]}, \left\{ \mathbf{x}^{[l']}_{d,1:T_{l'}} \right\}_{l'=1}^{L-1}, s_{d,t_{L}}^{[L]}, \Sigma^{[L]}_{\textbf{w}_{d}^{[L]}} \right) \\
    \textbf{y}_{d,t_{l}}^{[l]} &\vert \textbf{x}_{d,t_{l}}^{[l]} \sim \mathcal{N}\left( \textbf{y}_{d,t_{l}}^{[l]} \mid \textbf{x}_{d,t_{l}}^{[l]}, \Sigma^{[l]}_{\mathbf{v}^{[l]}_{d}} \right), \quad \text{for} \, l = 1, \dots, L \\
    \textbf{s}_{d,t_{L}}^{[L]} &\vert \mathcal{S}_{d,t_{L}-1}^{[L]}, \pi_{d,t_{L}}^{[L]} \sim \text{Cat}\left( \textbf{s}_{d,t_{L}}^{[L]}  \vert  \mathcal{S}_{d,t_{L}-1}^{[L]}, \boldsymbol{\pi}_{d,t_{L}}^{[L]} \right) \\
    \boldsymbol{\pi}_{d,t_{L}}^{[L]} &\vert \boldsymbol{\alpha} \sim \text{Dir}\left( \boldsymbol{\pi}_{d,t_{L}}^{[L]} \vert  \boldsymbol{\alpha} \right)
\end{align*}
where \( \mathcal{S}_{d,t_{L}-1}^{[L]} \) denotes the set of indicators from \( t_{L} = 1,...,t_{L}-1 \). The state transition distributions for \( \textbf{x}_{d,t_{l}}^{[l]} \) and \( \textbf{x}_{d,t_{L}}^{[L]} \) are modeled as multivariate normal distributions, \( \mathcal{N} \). This choice is appropriate for capturing the continuous evolution of the states over time, allowing for Gaussian noise in the state transition process, represented by the known covariance matrices \( \Sigma^{[l]}_{\textbf{w}_{d}^{[l]}} \) and \( \Sigma^{[L]}_{\textbf{w}_{d}^{[L]}} \). The measurement model \( \textbf{y}_{d,t_{l}}^{[l]} \) given the state \( \textbf{x}_{d,t_{l}}^{[l]} \) is also modeled using a Gaussian distribution, reflecting the assumption that measurements are continuous and subject to Gaussian measurement noise represented by \( \Sigma^{[l]}_{\mathbf{v}^{[l]}_{d}} \). The model indicators \( \textbf{s}_{d,t_{L}}^{[L]} \) are drawn from a categorical distribution, \( \text{Cat} \), which captures the discrete nature of model selection, allowing the model to switch between different regimes effectively. The model probabilities \( \boldsymbol{\pi}_{d,t_{L}}^{[L]} \) are drawn from a Dirichlet distribution, \( \text{Dir} \), which serves as a conjugate prior for the categorical distribution of the model indicators \cite{gelman1995bayesian}. The Dirichlet distribution's conjugacy simplifies the inference process, allowing straightforward updates based on the counts of model selections while incorporating prior beliefs represented by the concentration parameters \( \boldsymbol{\alpha} \). The joint posterior we wish to sample from is given as follows:
\begin{align}
&p\left( \left\{\textbf{x}_{d,t_{l}}^{[l]}, s_{d,t_{L}}^{[L]}, \pi_{d,t_{L}}^{L,(m)} \right\}_{d=1,m=1,l=1,t_{l}=1}^{D,M,L,T_{l}} \mid \left\{ \textbf{y}_{d,t_{l}}^{[l]} \right\}_{d=1,l=1,t_{l}=1}^{D,L,T_{l}} , \boldsymbol{\alpha},  \left\{\Sigma_{\textbf{w}_{d,t_{l}}^{[l]}}, \Sigma_{\textbf{v}_{d,t_{l}}^{[l]}} \right\}_{d=1,l=1}^{D,L} \right) 
\\
&=\prod_{d=1}^{D} \prod_{m=1}^{M} \prod_{l=1}^{L} \prod_{t_{l}=1}^{T_{l}} p( \textbf{x}_{d,t_{l}}^{[l]}, s_{d,t_{L}}^{[L]}, \pi_{d,t_{L}}^{L,(m)} \mid \textbf{y}_{d,t_{l}}^{[l]}, \alpha_{m}, \Sigma_{\textbf{w}_{d,t_{l}}^{[l]}}, \Sigma_{\textbf{v}_{d,t_{l}}^{[l]}}) \\
&\propto \prod_{d=1}^{D} \prod_{l=1}^{L} \prod_{t_{l}=1}^{T_{l}} p( \textbf{x}_{d,t_{l}}^{[l]} \mid \textbf{x}_{d,t_{l}-1}^{[l]}, \left\{ \mathbf{x}^{[l']}_{d,1:T_{l'}} \right\}_{l'=1}^{l-1},  \left\{ \mathbf{x}^{[l']}_{d,t_{l'}-1} \right\}_{l'=l+1}^{L}\Sigma_{\textbf{w}_{d,t_{l}}^{[l]}}) p(\textbf{y}_{d,t_{l}}^{[l]} \mid \textbf{x}_{d,t_{l}}^{[l]}, \Sigma^{[l]}_{\mathbf{v}^{[l]}_{d}}) \nonumber \\
& p( \textbf{x}_{d,t_{L}}^{[L]} \mid \textbf{x}_{d,t_{L}-1}^{[L]}, \left\{ \mathbf{x}^{[l']}_{d,1:T_{l'}} \right\}_{l'=1}^{L-1}, s_{d,t_{L}}^{[L]}, \Sigma_{\textbf{w}_{d,t_{L}}^{[L]}})   p( \textbf{s}_{d,t_{L}}^{[L]} \mid \mathcal{S}_{d,t_{L}-1}^{[L]}, \pi_{d,t_{L}}^{[L],(m)}) p(\boldsymbol{\pi}_{d,t_{L}}^{[L]} \mid \boldsymbol{\alpha})
\end{align}
With this hierarchical structure established, we can now proceed to describe the inference process, which leverages sequential Monte Carlo (SMC) to estimate the posterior distributions of the latent states and model indicators.

\subsection{Inference via Multiscale Sequential Monte Carlo (SMC)}
For inference, we leverage a multiscale Sequential Monte Carlo (SMC) approach to account for the nested structure of the dynamics. This allows us to propagate particles across different scales, handling both coarse and fine-scale dynamics simultaneously. The key novelty lies in the ability of the SMC algorithm to handle nested nonlinear dynamics at each level of the model, where the states at finer time scales feed into the coarser scales and vice versa. To simplify the weight calculation, the algorithm uses the joint prior as the proposal distribution \cite{arulampalam2002tutorial} as
\begin{align}
    &p(\textbf{x}_{d,t_{l}}^{[l]}, \textbf{x}_{d,t_{l}-1}^{[l]},  \left\{ \mathbf{x}^{[l']}_{d,1:T_{l'}} \right\}_{l'=1}^{l-1},  \left\{ \mathbf{x}^{[l']}_{d,t_{l'}-1} \right\}_{l'=l+1}^{L}, s_{d,t_{L}}^{[L]} = m, \mathcal{S}_{d,t_{L}-1}^{[L]}, \pi_{d,t_{L}}^{[L],(m)}, \alpha_{m})\\
    &\propto p(\textbf{x}_{d,t_{L}}^{[L]} \vert \textbf{x}_{d,t_{L}-1}^{[L]},  \left\{ \mathbf{x}^{[l'], (i)}_{d,1:T_{l'}} \right\}_{l'=1}^{L-1}, s_{d,t_{L}}^{[L]} = m,\Sigma_{\textbf{w}_{d,t_{L}}^{[L]}} ) \\
    &p(\textbf{x}_{d,t_{l}}^{[l]} \vert \textbf{x}_{d,t_{l}-1}^{[l]},  \left\{ \mathbf{x}^{[l']}_{d,1:T_{l'}} \right\}_{l'=1}^{l-1},  \left\{ \mathbf{x}^{[l']}_{d,t_{l'}-1} \right\}_{l'=l+1}^{L}, \Sigma_{\textbf{w}_{d,t_{l}}^{[l]}})\\
    & p( \textbf{s}_{d,t_{L}}^{[L]} \vert \mathcal{S}_{d,t_{L}-1}^{[L]}, \pi_{d,t_{L}}^{[L],(m)}) p(\boldsymbol{\pi}_{d,t_{L}}^{[L]} \vert \boldsymbol{\alpha})
\end{align}
For each $l = 1, \dots, L-1$, where each finer scale $l$ is nested within the coarser time scale $l+1$, and for each particle $i=1,...,N_{s}$ the states $\textbf{x}_{d,t_{l}}^{[l], (i)}$ are sampled from $p(\textbf{x}_{d,t_{l}}^{[l], (i)} \vert \textbf{x}_{d,t_{l}-1}^{[l],(i)},  \left\{ \mathbf{x}^{[l'], (i)}_{d,1:T_{l'}} \right\}_{l'=1}^{l-1},  \left\{ \mathbf{x}^{[l', (i)]}_{d,t_{l'}} \right\}_{l'=l+1}^{L}, \Sigma_{\textbf{w}_{d,t_{l}}^{[l]}})$ for $t_{l} = 1, \dots, T_l$ using the standard SMC algorithm. The resampled particles from each time step $t_{l}$ at time scale $l$ serve as input to the particle filter at the coarser time scale $l+1$ for time step $t_{l}+1$. Similarly, the output of the coarser time scale $l+1$ particle filter at time $t_{l+1}$ becomes an input into the finer time scale PF at the next iteration $t_{l-1}+1$. Then, for the coarsest time scale $L$ the probabilities $\pi_{d,t_{L}}^{[L],(m),(i)}$ for selecting each model $m=1,...,M$ are sampled from the following Dirichlet distribution:
\begin{equation}
    \boldsymbol{\pi}_{d,t_{L}}^{[L], (i)} \vert \textbf{s}_{d,1:t_{L}-1}^{[L], (i)}, \boldsymbol{\alpha} \sim \text{Dir}(\alpha_{1} + c_{1:t_{L}-1}^{[L],(1), (i)},...,\alpha_{m} + c_{1:t_{L}-1}^{[L],(M), (i)})
\end{equation}
where $c_{1:t_{L}-1}^{[L],(M), (i)}$ is the number of times model $m$ has been selected up to time $t_{L}-1$ and $\alpha_{m}$ is the Dirichlet distribution concentration parameter for model $m$. Next, the model indicator $ s_{d,t_{L}}^{[L], (i)}$ is sampled from
\begin{equation}
\begin{aligned}
&p( s_{d,t_{L}}^{[L], (i)} = m \vert \mathcal{S}_{d,t_{L}-1}^{[L],(i)}, \textbf{x}_{d,t_{L}}^{L, (i)},  \left\{ \mathbf{x}^{[l', (i)]}_{d,1:T_{l'}} \right\}_{l'=1}^{l-1}, \pi_{d,t_{L}}^{[L], (m), (i)}, \textbf{y}_{d,t_{L}}^{[L]}) \\
&\propto p(s_{d,t_{L}}^{[L], (i)} = m \vert  \mathcal{S}_{d,t_{L}-1}^{[L],(i)}, \pi_{d,t_{L}}^{[L], (m), (i)}) p(\textbf{x}_{d,t_{L}}^{[L], (i)} \vert \textbf{x}_{d,t_{L}-1}^{[L],(i)},  \left\{ \mathbf{x}^{[l'],(i)}_{d,1:T_{l'}} \right\}_{l'=1}^{l-1}, s_{d,t_{L}}^{[L],(i)}= m, \Sigma_{\textbf{w}_{d,t_{L}}^{[L]}}) \\
&p(\textbf{y}_{d,t_{l}}^{[l],(i)} \vert \textbf{x}_{d,t_{l}}^{[l],(i)}, \Sigma^{[l]}_{\mathbf{v}^{[l]}_{d}}) 
\end{aligned}
\end{equation}
It is selected by first computing the prior probability $p(s_{d,t_{L}}^{[L], (i)} = m \vert \mathcal{S}_{d,t_{L}-1}^{[L],(i)}, \pi_{d,t_{L}}^{[L], (m), (i)})$, which reflects how frequently the model $m$ has been chosen in the past, weighted by the Dirichlet-distributed parameters $\pi_{d,t_{L}}^{[L],(m), (i)}$. This prior is then combined with the transition density and measurement likelihood to form a posterior distribution. From this posterior, the model indicator $s_{d,t_{L}}^{[L], (i)}$ is sampled, determining the most probable model $m$ for the current time step based on past selections and the current state and measurement dynamics. More specifically, for each model $m$, a new state $\textbf{x}_{d,t_{L}}^{L, (i)*}$ is sampled from 
\begin{equation}
\label{eqn:samplestate}
\textbf{x}_{d,t_{L}}^{L, (i)*} = f_L\left( \mathbf{A}^{[L]}_{t_L}\mathbf{x}^{[L], (i)}_{d,t_L-1}, \left\{ \mathbf{x}^{[l', (i)]}_{d,1:T_{l'}} \right\}_{l'=1}^{L-1}, s_{d,t_{L}}^{[L], (i)} = m, \sum_{d' \neq d} B^{[L]}_{d,d'} \mathbf{x}^{[L], (i)}_{d',t_L-1} \right) +  \mathbf{w}^{[L]}_{d,t_{L}-1}
\end{equation}
and the corresponding transition densities are computed for each model
\[
p(\textbf{x}_{d,t_{L}}^{[L], (i)*} \vert \textbf{x}_{d,t_{L}-1}^{[L], (i)},  \left\{ \mathbf{x}^{[l'], (i)}_{d,1:T_{l'}} \right\}_{l'=1}^{l-1}, s_{d,t_{L}}^{[L], (i)} = m, \Sigma_{\textbf{w}_{d,t_{L}}^{[L]}}).
\]
Then the likelihoods for each model are computed via
\[
p(\textbf{y}_{d,t_{l}}^{[l],(i)} \vert \textbf{x}_{d,t_{L}}^{L, (i)*}, \Sigma^{[l]}_{\mathbf{v}^{[l]}_{d}}).
\]
Given the selected model indicator, $s_{d,t_{L}}^{[L], (i)} = m$, a new state $\textbf{x}_{d,t_{L}}^{L, (i)}$ is again sampled from Equation \ref{eqn:samplestate}. The particle weights are computed and then the states and indicators are resampled. These steps are summarized in Algorithm 1. 
\newpage
\begin{algorithm}[H]
\caption{Multiscale Particle Filter}
\begin{algorithmic}[1]
\State \textbf{Input:} Number of particles $N_s$, time scales $L$, measurements $\mathbf{y}_{d,t_l}$, transition functions $f_l$, Dirichlet concentration parameters $\alpha_m$, interaction matrices $\mathbf{A}, \mathbf{B}$, noise covariance matrices $\Sigma_{\mathbf{w}}, \Sigma_{\mathbf{v}}$
\State \textbf{Initialize:} 
    \For{each individual $d$}
        \For{each time scale $l = 1, \dots, L$}
            \State Initialize particles $\mathbf{x}_{d,0}^{[l], (i)}$ for $i = 1, \dots, N_s$
            \State Set initial weights $w_{d,0}^{[l], (i)} = \frac{1}{N_s}$ for all particles
        \EndFor
    \EndFor
\State
\For{each individual $d$}
\For{each time scale $l = 1, \dots, L-1$}
    \For{each time step $t_{l} = 1, \dots, T_{l}$}
            \For{each particle $i = 1, \dots, N_s$}
                \State Sample $\mathbf{x}_{d,t_{l}}^{[l], (i)}$ from
                \[
                \mathbf{x}^{[l], (i)*}_{d,t_l} = f_l\left( \mathbf{A}^{[l]}_{t_l}\mathbf{x}^{[l], (i)}_{d,t_l-1}, \left\{ \mathbf{x}^{[l'], (i)}_{d,1:T_{l'}} \right\}_{l'=1}^{l-1},  \left\{ \mathbf{x}^{[l'], (i)}_{d,t_{l'}-1} \right\}_{l'=l+1}^{L} \right) +  \mathbf{w}^{[l]}_{d,t_{l}-1}
                \]
        \State Update particle weights using the previous weights and the computed likelihood:
\[
w_{d,t_{l}}^{[l], (i)} \propto w_{d,t_{l}-1}^{[l], (i)} \cdot p(\mathbf{y}_{d,t_{l}}^{[l]} \mid \mathbf{x}_{d,t_{l}}^{[l], (i)}, \Sigma^{[L]}_{\mathbf{v}^{[L]}_{d}})
\]
          \EndFor
        \State Resample particles at time step $t_l$, where the resampled particles serve as input to the coarser time scale $l+1$ at the next iteration.
        \EndFor
    \EndFor

\For{each time step $t_{L} = 1, \dots, T_{L}$ at the coarsest time scale $L$}
        \For{each particle $i = 1, \dots, N_s$}
            \State Sample model probabilities $\boldsymbol{\pi}_{d,t_{L}}^{[L], (i)} \sim \text{Dir}(\alpha_{1} + c_{1:t_{L}-1}^{[L],(1), (i)}, \dots, \alpha_{M} + c_{1:t_{L}-1}^{[L],(M), (i)})$
            \State Sample model indicator $s_{d,t_{L}}^{[L], (i)}$ from
            \begin{align*}
            &p( s_{d,t_{L}}^{[L], (i)} = m \vert \mathcal{S}_{d,t_{L}-1}^{[L],(i)}, \textbf{x}_{d,t_{L}}^{[L], (i)},  \left\{ \mathbf{x}^{[l'], (i)]}_{d,1:T_{l'}} \right\}_{l'=1}^{l-1}, \pi_{d,t_{L}}^{[L], (m), (i)}, \textbf{y}_{d,t_{L}}^{[L]}) \\
&\propto p(s_{d,t_{L}}^{[L], (i)} = m \vert  \mathcal{S}_{d,t_{L}-1}^{[L],(i)}, \pi_{d,t_{L}}^{[L], (m), (i)}) p(\textbf{x}_{d,t_{L}}^{[L], (i)} \vert \textbf{x}_{d,t_{L}-1}^{[L],(i)},  \left\{ \mathbf{x}^{[l'],(i)}_{d,1:T_{l'}} \right\}_{l'=1}^{l-1}, s_{d,t_{L}}^{[L],(i)}= m, \Sigma_{\textbf{w}_{d,t_{L}}^{[L],(i)}}) \\
&p(\textbf{y}_{d,t_{l}}^{[l]} \vert \textbf{x}_{d,t_{l}}^{[l],(i)}, \Sigma^{[l]}_{\mathbf{v}^{[l]}_{d}}) 
            \end{align*}
    
            \State Sample new state $\mathbf{x}_{d,t_{L}}^{L, (i)}$ from:
            \[
            \mathbf{x}_{d,t_{L}}^{L, (i)} = f_L\left( \mathbf{A}^{[L]}_{t_L}\mathbf{x}^{[L]}_{d,t_L-1}, \left\{ \mathbf{x}^{[l'], (i)}_{d,1:T_{l'}} \right\}_{l'=1}^{L-1}, s_{d,t_{L}}^{[L], (i)} = m, \sum_{d' \neq d} \mathbf{B}^{[L]}_{d,d'} \mathbf{x}^{[L], (i)}_{d',t_L-1} \right) + \mathbf{w}^{[L]}_{d,t_{L}-1}
            \]
           \State Compute the likelihood of the current observation:
    \State Update particle weights using the previous weights and the computed likelihood:
\[
w_{d,t_{L}}^{[L], (i)} \propto w_{d,t_{L}-1}^{[L], (i)} \cdot p(\mathbf{y}_{d,t_{L}}^{[L]} \mid \mathbf{x}_{d,t_{L}}^{L, (i)}, \Sigma^{[L]}_{\mathbf{v}^{[L]}_{d}})
\]
  \EndFor
\EndFor
\State Resample particles based on updated weights

    \EndFor
\end{algorithmic}
\end{algorithm}

\section{Simulations}
In this study, we implement a multiscale switching model to evaluate the dynamics of state transitions across multiple time scales, specifically for $L=2$. The model simulates a scenario involving three state dimensions for $l=2$, three state dimensions for $l=1$, and six individuals, with particle filtering employed to estimate hidden states. The simulation is set up with 100 time steps for $l=2$ and 50 time steps for $l=1$ per $l=2$ step. The initial conditions for the particles are determined by three different starting states for each individual, specifically set to $0.2$, $0.5$, and $0.7$. We use $N_{s} = 1,000$ particles. For all simulations, the dimensions of the states for $l=1$ and $l=2$ are $N_{L} = 3$.

\subsection{Simulation 1}
The true dynamics for $l=2$ follow two distinct models over time. In the first third of the simulation (time steps $t = 1$ to $t = 33$), the true dynamics are governed by model $m=0$, characterized by strong oscillations:
\begin{equation}
\mathbf{x}_{d,t_2}^{[2]} = 3 \sin(\mathbf{x}_{d,t_2-1}^{[2]} + \frac{\pi}{4}) + \bigg(\frac{\sum_{t_{1}=1}^{T_{1}} w_{t_{1}} \mathbf{x}_{d,t_{1}}^{[1]}}{\sum_{t_{1}=1}^{T_{1}} w_{t_{1}}}\bigg) + 0.5  \sum_{d' \neq d} B^{[2]}_{d,d'} \mathbf{x}^{[2]}_{d',t_{2}-1} + 0.3\mathbf{A}^{[2]}\mathbf{x}^{[2]}_{d,t_2-1}.
\end{equation}
In the second third of the simulation (time steps $t = 34$ to $t = 66$), the dynamics switch to model $m=1$ as follows:
\begin{align}
&\mathbf{x}_{d,t_2}^{[2]} = 2 \cos(1.2 \cdot \mathbf{x}_{d,t_2-1}^{[2]}) \cdot \exp(-0.05 \cdot \mathbf{x}_{d,t_2-1}^{[2]}) + \bigg(\frac{\sum_{t_{1}=1}^{T_{1}} w_{t_{1}} \mathbf{x}_{d,t_{1}}^{[1]}}{\sum_{t_{1}=1}^{T_{1}} w_{t_{1}}}\bigg) \\
&+  \sum_{d' \neq d} B^{[2]}_{d,d'} \mathbf{x}^{[2]}_{d',t_2-1 } + 0.5\mathbf{A}^{[2]}\mathbf{x}^{[2]}_{d,t_2-1}.
\end{align}
The dynamics then revert to model $m=0$ for the final third of the simulation (time steps $t = 67$ to $t = 100$). For time scale $l=1$, the dynamics are given by the following equation:
\begin{equation}
\mathbf{x}_{d,t_1}^{[1]} = \cos(1 +\mathbf{A}^{[1]}\mathbf{x}^{[1]}_{d,t_{1}-1}) + 0.6\mathbf{x}_{d,t_2-1}^{[2]} 
\end{equation}
Noise is introduced into the models through various covariance matrices. Specifically, the transition noise for $l=2$ for each individual $d$ is defined as follows: 
$\Sigma_{\mathbf{w}_{1,t_2}^{[2]}} = 0.5 \, \mathbf{I}_{N_2}, \, \Sigma_{\mathbf{w}_{2,t_2}^{[2]}} = 0.4 \, \mathbf{I}_{N_2}, \, \Sigma_{\mathbf{w}_{3,t_2}^{[2]}} = 0.5 \, \mathbf{I}_{N_2}, \, \Sigma_{\mathbf{w}_{4,t_2}^{[2]}} = 0.7 \, \mathbf{I}_{N_2}, \, \Sigma_{\mathbf{w}_{5,t_2}^{[2]}} = 0.3 \, \mathbf{I}_{N_2}, \, \Sigma_{\mathbf{w}_{6,t_2}^{[2]}} = 0.4 \, \mathbf{I}_{N_2}.$ In contrast, the transition noise for $l=1$ is characterized by a lower variance, modeled as
$\Sigma_{\mathbf{w}_{d}^{[1]}} = 0.05 \, \mathbf{I}_{N_1}$.
Similarly, the measurement noise for the time scales is specified as follows: for $l=2$, the covariance is given by $\Sigma_{\mathbf{v}_{d_{t_{2}}}^{[2]}} = 0.02 \, \mathbf{I}_{N_2}$, while for $l=1$, the measurement noise is represented as $\Sigma_{\mathbf{v}_{d_{1}}^{[1]}} = 0.03 \, \mathbf{I}_{N_1}$, which is consistent across all individuals. The interaction matrix $ \mathbf{B}^{[2]} $ is initialized as an identity matrix, with one randomly chosen off-diagonal entry set to 1, capturing basic interactions among individuals. Additionally, the adjacency matrix $ \mathbf{A}^{[1]} $ is randomly generated, consisting of zeros and ones to define the interaction structure between dimensions. 

\subsubsection{Results and Discussion}
\begin{figure}[htbp]
    \centering
    \subfigure[Individual 1: State Estimate with $\Sigma_{\mathbf{w}_{1,t_2}^{[2]}} = 0.5 \, \mathbf{I}_{N_2}$]{
        \includegraphics[width=0.4\textwidth]{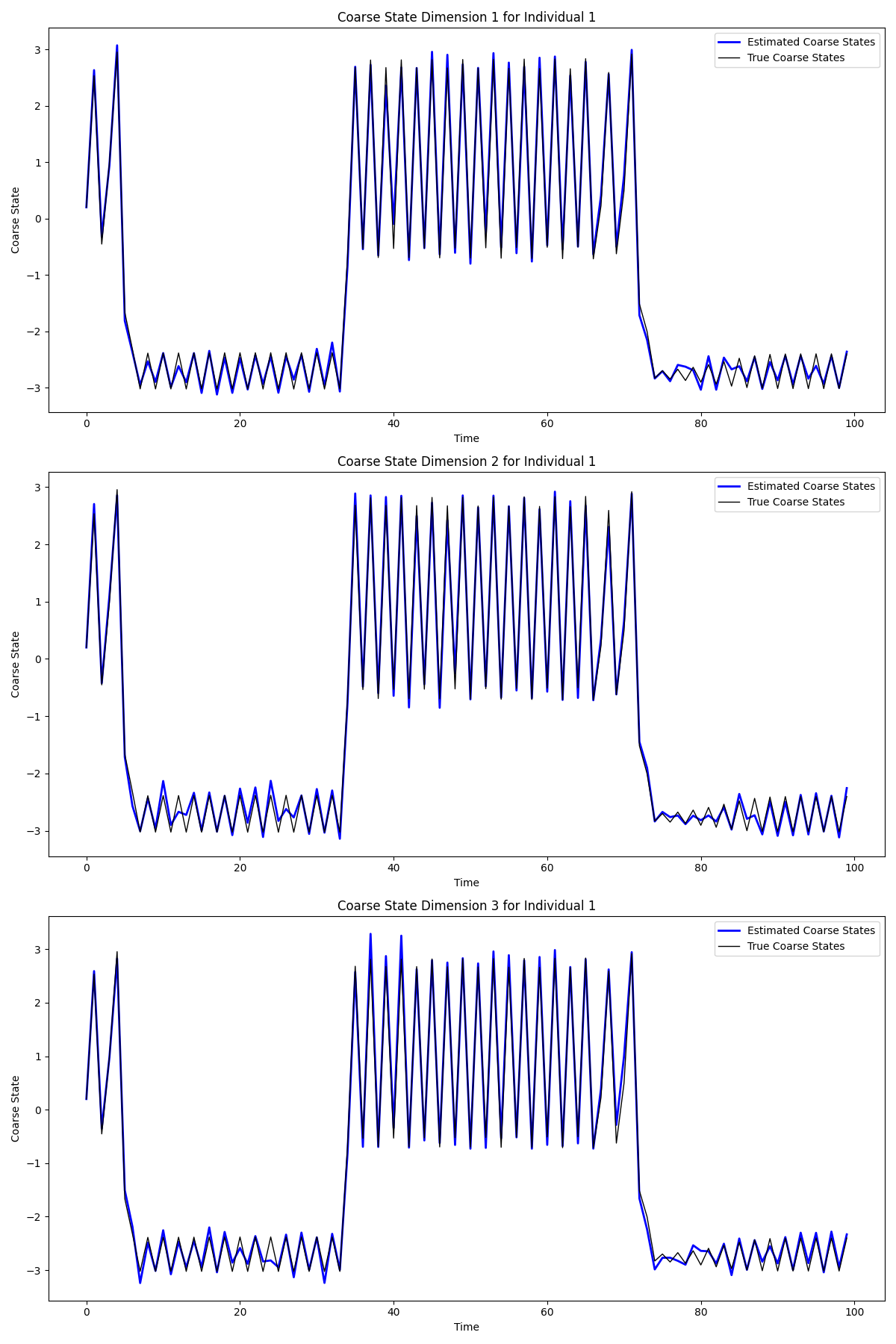}
        \label{fig:individual1_state}
    }
    \subfigure[Individual 1: True vs. Estimated Indicator with $\Sigma_{\mathbf{w}_{1,t_2}^{[2]}} = 0.5 \, \mathbf{I}_{N_2}$]{
        \includegraphics[width=0.8\textwidth]{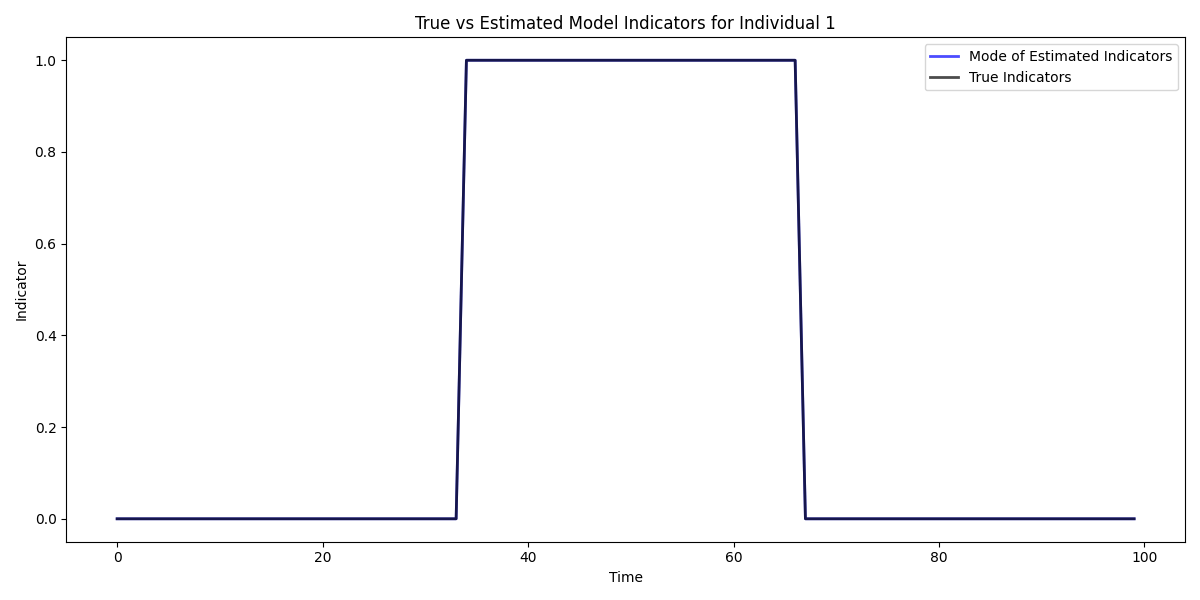}
        \label{fig:individual1_indicator}
    }
    \captionsetup{width=0.9\textwidth}
    \caption{Comparison of true states and estimated states for Individual 1, along with the estimated indicator. In the indicator plot, blue indicates the estimated model indicator while black represents the true model indicator.}
    \label{fig:true_vs_estimated_states_individual1}
\end{figure}

\begin{figure}[htbp]
    \centering
    \subfigure[Individual 2: State Estimate with $\Sigma_{\mathbf{w}_{2,t_2}^{[2]}} = 0.4 \, \mathbf{I}_{N_2}$]{
        \includegraphics[width=0.45\textwidth]{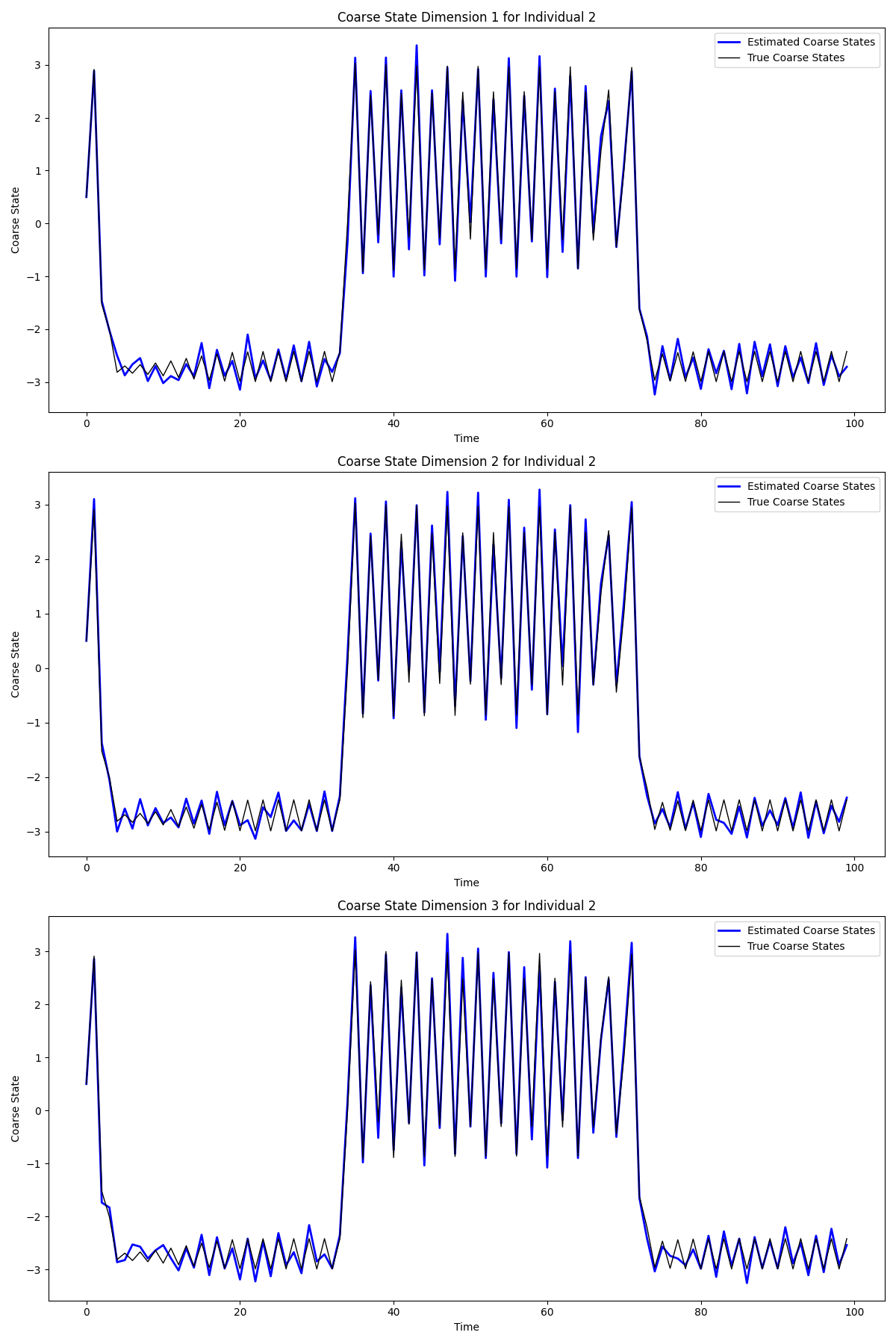}
        \label{fig:individual2_state}
    }
    \subfigure[Individual 2: True vs. Estimated Indicator with $\Sigma_{\mathbf{w}_{2,t_2}^{[2]}} = 0.4 \, \mathbf{I}_{N_2}$]{
        \includegraphics[width=0.8\textwidth]{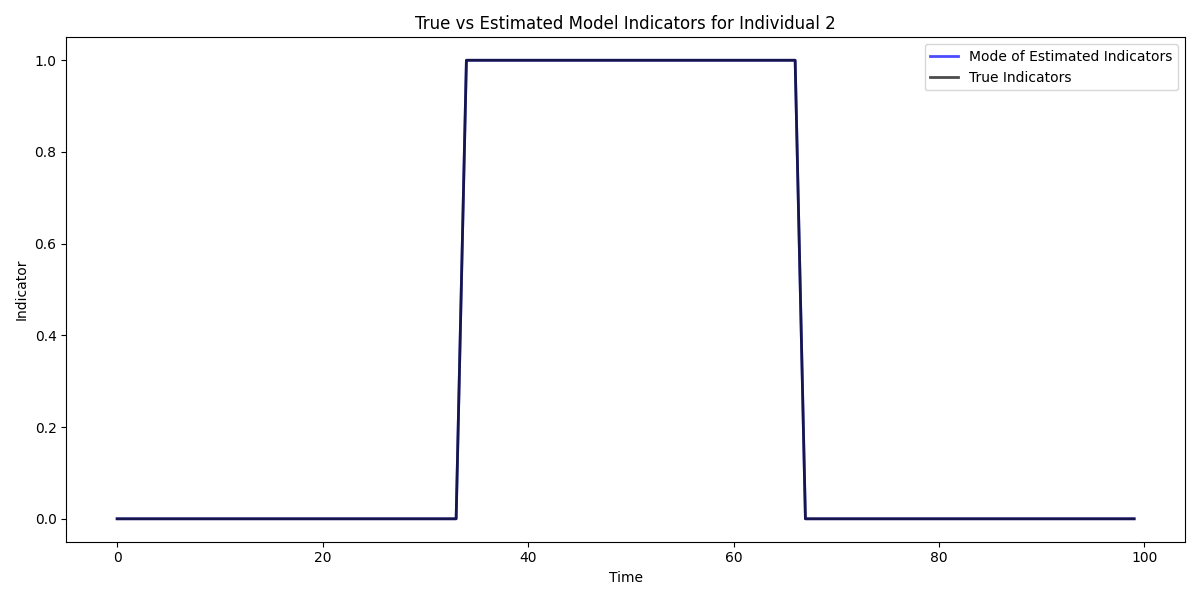}
        \label{fig:individual2_indicator}
    }
    \captionsetup{width=0.9\textwidth}
    \caption{Comparison of true states and estimated states for Individual 2, along with the estimated indicator. In the indicator plot, blue indicates the estimated model indicator while black represents the true model indicator.}
    \label{fig:true_vs_estimated_states_individual2}
\end{figure}

\begin{figure}[htbp]
    \centering
    \subfigure[Individual 3: State Estimate with $\Sigma_{\mathbf{w}_{3,t_2}^{[2]}} = 0.5 \, \mathbf{I}_{N_2}$]{
        \includegraphics[width=0.45\textwidth]{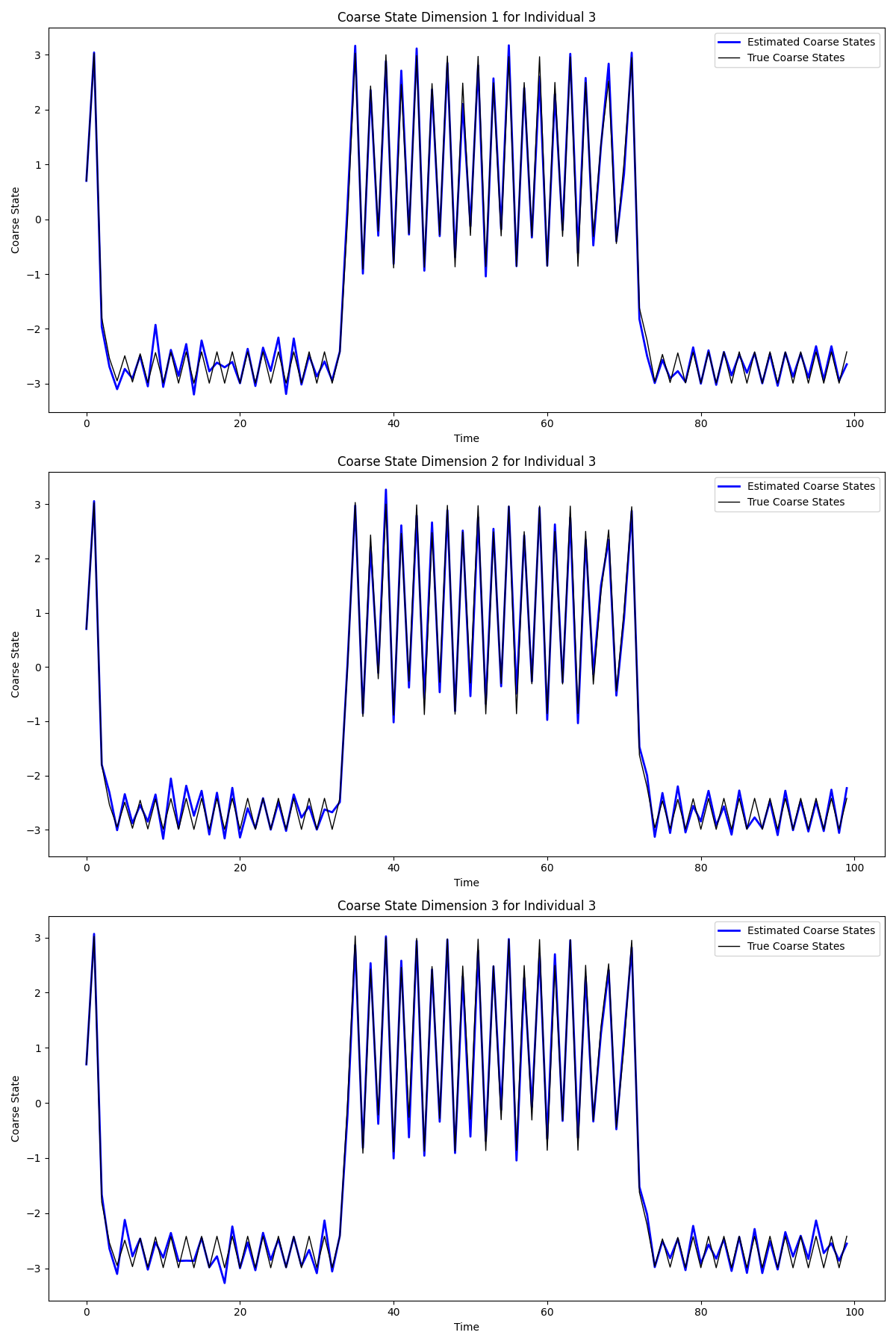}
        \label{fig:individual3_state}
    }
    \subfigure[Individual 3: True vs. Estimated Indicator with $\Sigma_{\mathbf{w}_{3,t_2}^{[2]}} = 0.5 \, \mathbf{I}_{N_2}$]{
        \includegraphics[width=0.8\textwidth]{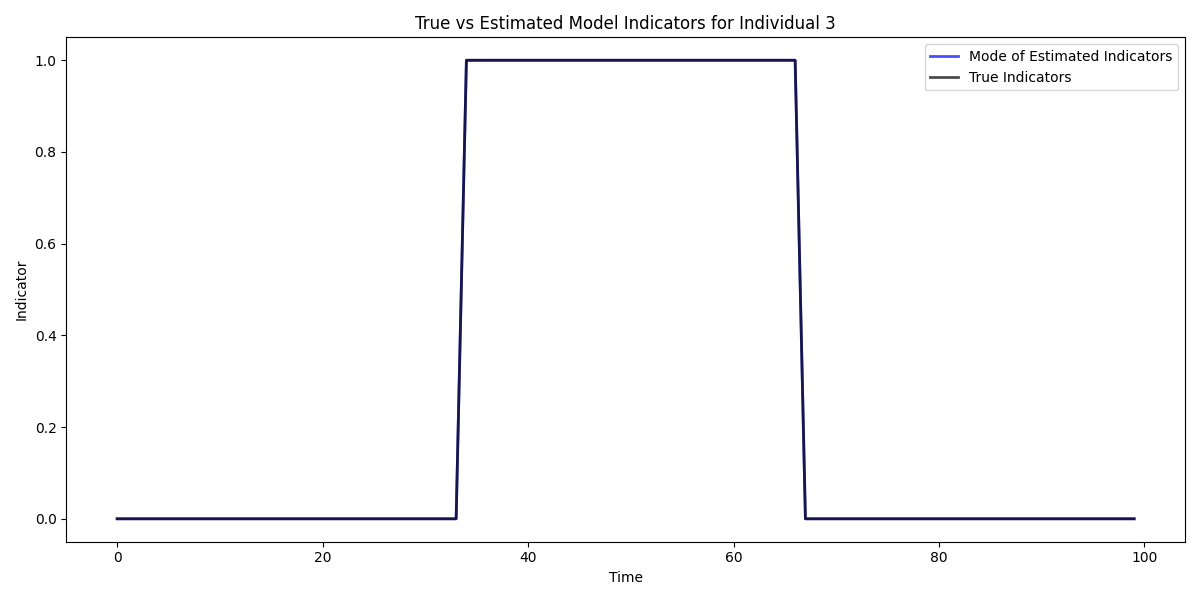}
        \label{fig:individual3_indicator}
    }
    \captionsetup{width=0.9\textwidth}
    \caption{Comparison of true states and estimated states for Individual 3, along with the estimated indicator. In the indicator plot, blue indicates the estimated model indicator while black represents the true model indicator.}
    \label{fig:true_vs_estimated_states_individual3}
\end{figure}

\begin{figure}[htbp]
    \centering
    \subfigure[Individual 4: State Estimate with $\Sigma_{\mathbf{w}_{4,t_2}^{[2]}} = 0.7 \, \mathbf{I}_{N_2}$]{
        \includegraphics[width=0.45\textwidth]{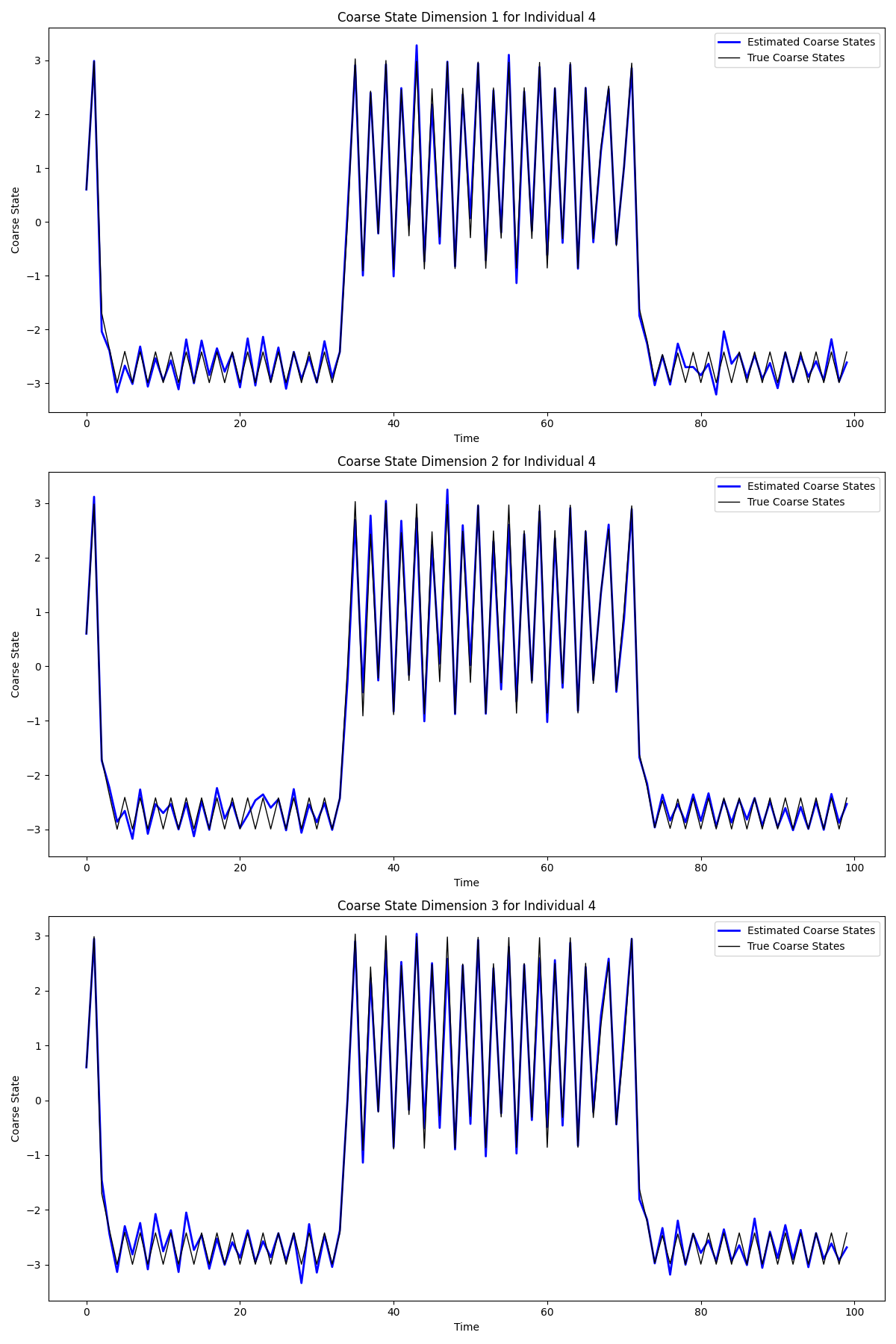}
        \label{fig:individual4_state}
    }
    \subfigure[Individual 4: True vs. Estimated Indicator with $\Sigma_{\mathbf{w}_{4,t_2}^{[2]}} = 0.7 \, \mathbf{I}_{N_2}$]{
        \includegraphics[width=0.8\textwidth]{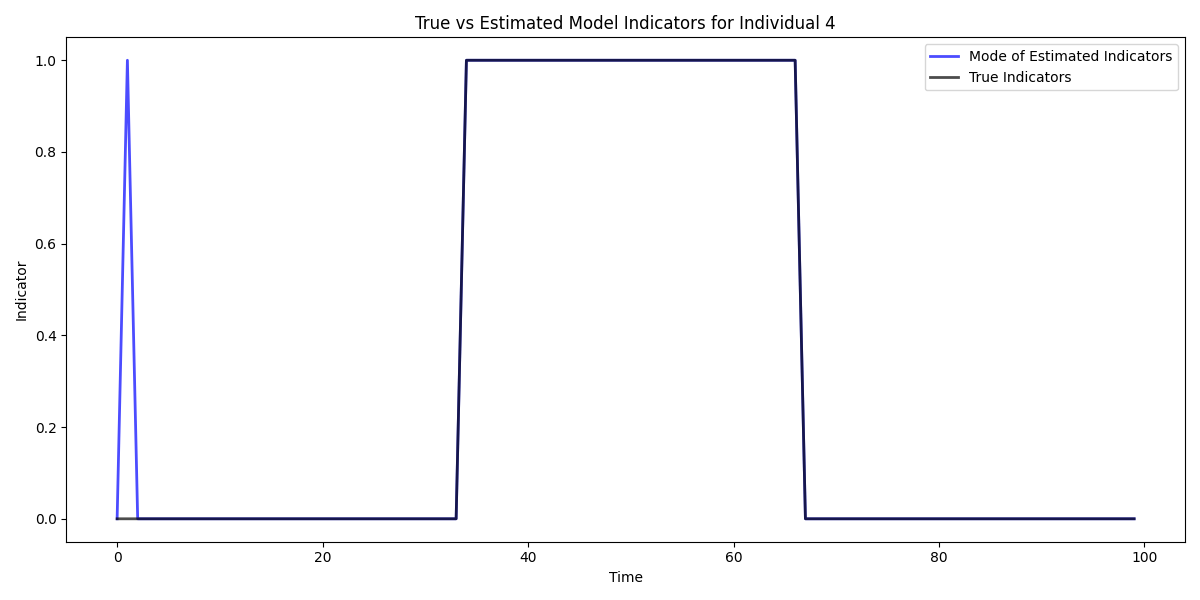}
        \label{fig:individual4_indicator}
    }
    \captionsetup{width=0.9\textwidth}
    \caption{Comparison of true states and estimated states for Individual 4, along with the estimated indicator. In the indicator plot, blue indicates the estimated model indicator while black represents the true model indicator.}
    \label{fig:true_vs_estimated_states_individual4}
\end{figure}

\begin{figure}[htbp]
    \centering
    \subfigure[Individual 5: State Estimate with $\Sigma_{\mathbf{w}_{5,t_2}^{[2]}} = 0.3 \, \mathbf{I}_{N_2}$]{
        \includegraphics[width=0.45\textwidth]{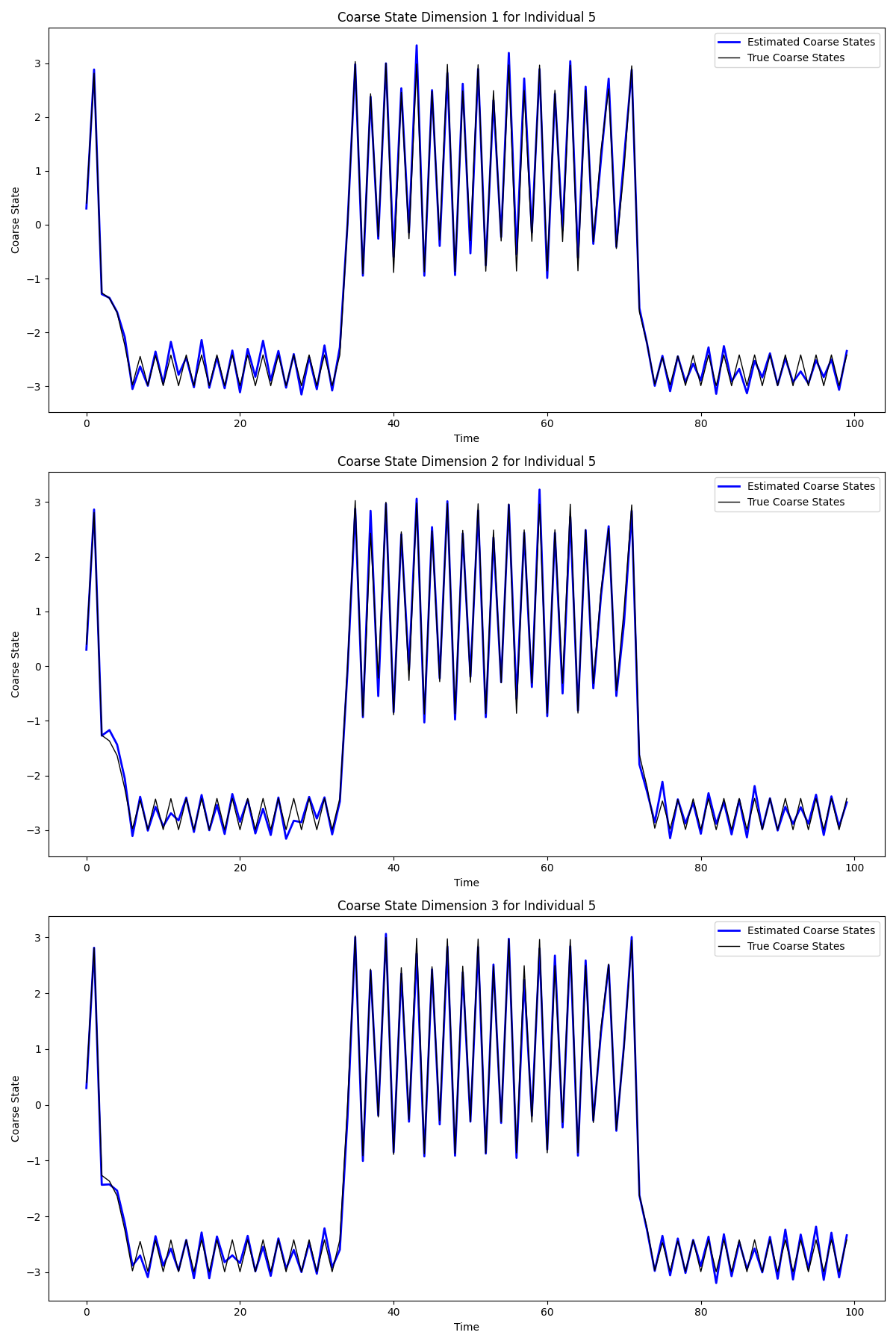}
        \label{fig:individual5_state}
    }
    \subfigure[Individual 5: True vs. Estimated Indicator with $\Sigma_{\mathbf{w}_{5,t_2}^{[2]}} = 0.3 \, \mathbf{I}_{N_2}$]{
        \includegraphics[width=0.8\textwidth]{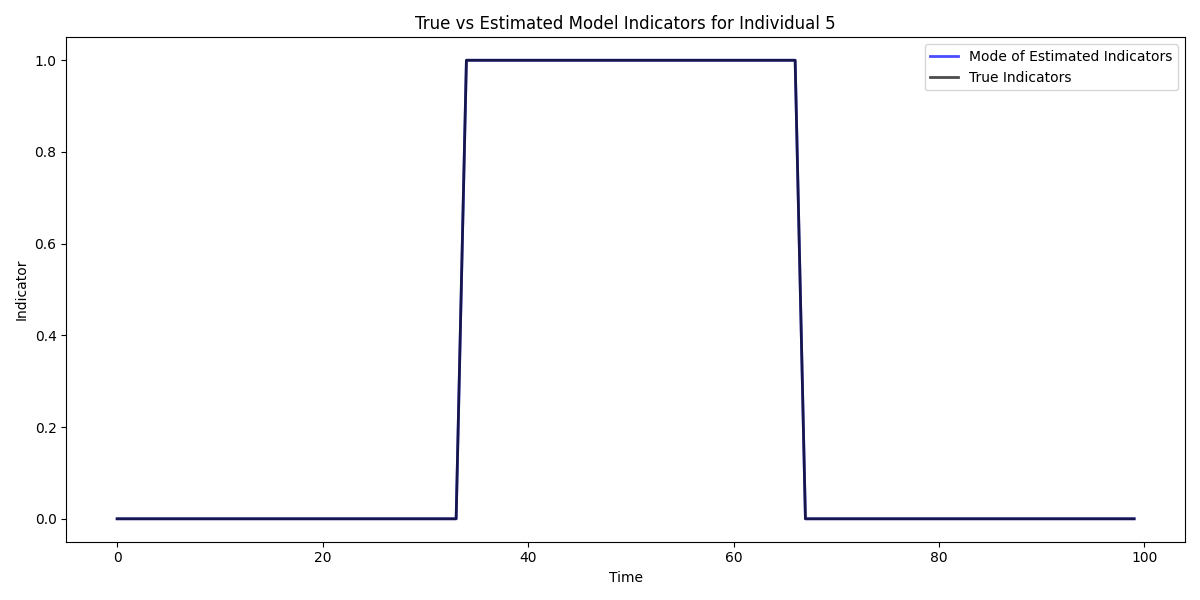}
        \label{fig:individual5_indicator}
    }
    \captionsetup{width=0.9\textwidth}
    \caption{Comparison of true states and estimated states for Individual 5, along with the estimated indicator. In the indicator plot, blue indicates the estimated model indicator while black represents the true model indicator.}
    \label{fig:true_vs_estimated_states_individual5}
\end{figure}

\begin{figure}[htbp]
    \centering
    \subfigure[Individual 6: State Estimate with $\Sigma_{\mathbf{w}_{6,t_2}^{[2]}} = 0.4 \, \mathbf{I}_{N_2}$]{
        \includegraphics[width=0.45\textwidth]{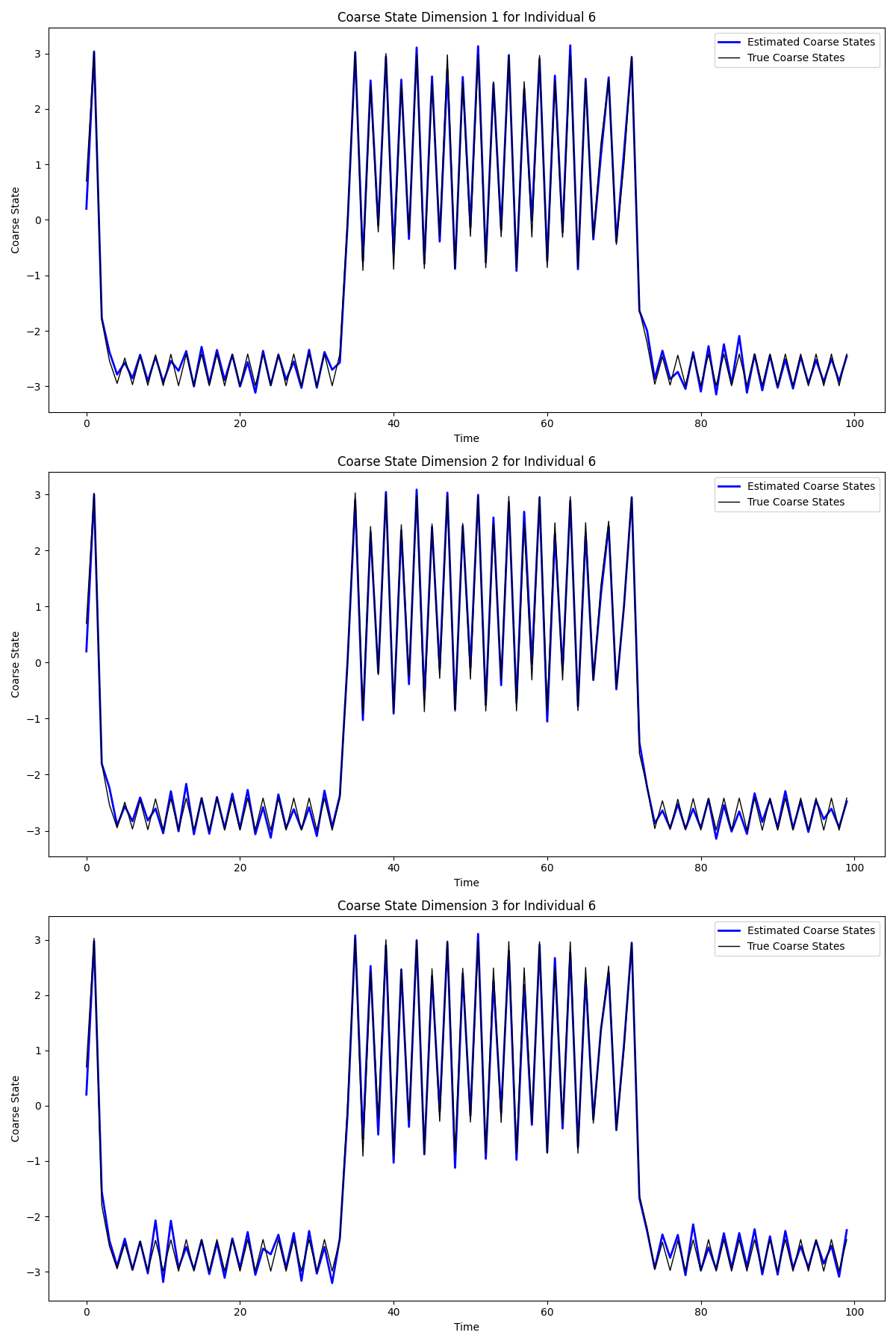}
        \label{fig:individual6_state}
    }
    \subfigure[Individual 6: True vs. Estimated Indicator with $\Sigma_{\mathbf{w}_{6,t_2}^{[2]}} = 0.4 \, \mathbf{I}_{N_2}$]{
        \includegraphics[width=0.8\textwidth]{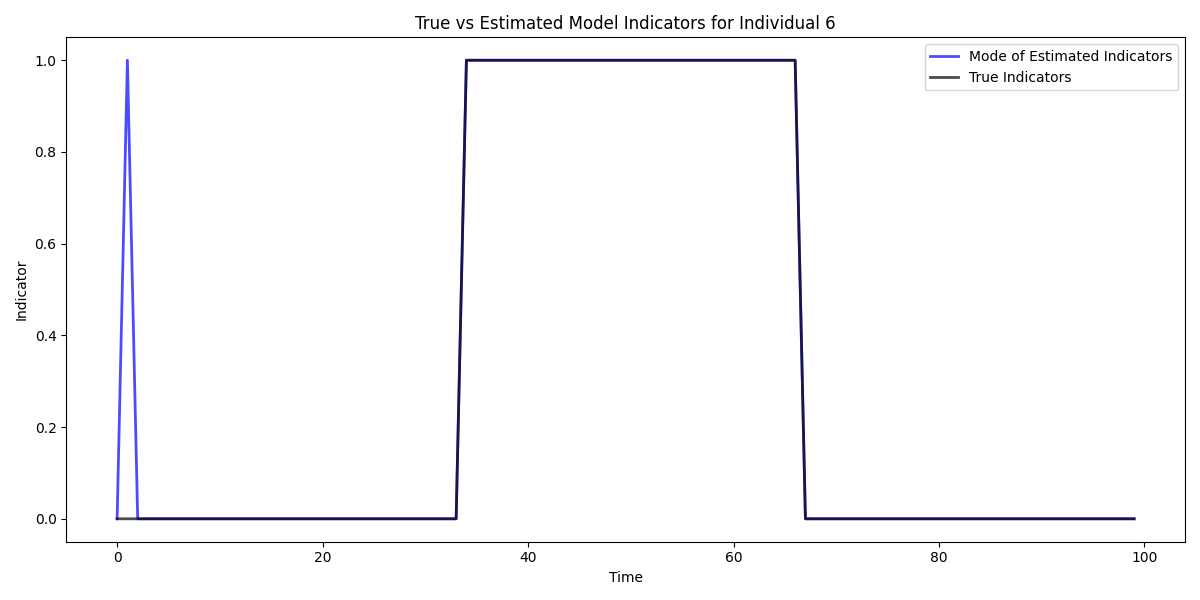}
        \label{fig:individual6_indicator}
    }
    \captionsetup{width=0.9\textwidth}
    \caption{Comparison of true states and estimated states for Individual 6, along with the estimated indicator. In the indicator plot, blue indicates the estimated model indicator while black represents the true model indicator.}
    \label{fig:true_vs_estimated_states_individual6}
\end{figure}

For all six individuals (Figures \ref{fig:true_vs_estimated_states_individual1} through \ref{fig:true_vs_estimated_states_individual6}), the estimated coarse states generally track the true states well throughout the time series, including during the dynamic transitions between states. In general, our algorithm effectively captures the overall shape and frequency of these oscillations but tends to overshoot or undershoot in some cases, particularly in the first and last third of the time series. Despite these discrepancies, the transitions between different state regimes are well captured, demonstrating the model's capacity to follow the underlying dynamics. The variations in the noise do not have a particularly significant effect on the estimation accuracy. The robustness in capturing the transitions between states highlights the strength of our algorithm, particularly in the more dynamic phases of the time series ($t_{2} = 34,...,66$). 

The plots for the true vs. estimated model indicators (Figures \ref{fig:individual1_indicator} through \ref{fig:individual6_indicator}) demonstrate that the switching between model indicators is captured accurately and without delay for all individuals. This is particularly significant as the correct identification of model transitions is crucial for ensuring that the appropriate model is selected at the correct time during the filtering process. The immediate capture of these transitions allows the filter to adapt to changing dynamics swiftly, thereby improving the overall accuracy of state estimation. This timely and accurate switching is especially beneficial in scenarios with nonlinear and complex dynamics, as any delay in detecting a model switch could propagate errors into the state estimates. For some individuals (e.g., Figures \ref{fig:individual4_indicator} and \ref{fig:individual6_indicator}), there is a slight misalignment at the very beginning of the time series, where the estimated indicator briefly spikes before settling into the correct model. This suggests a potential sensitivity to initial conditions that could be refined with additional prior information or smoothing during initialization. Overall, the accurate capture of transitions for the rest of the time series reflects the robustness of the learning algorithm and demonstrates that it is functioning as intended. The ability to consistently and promptly detect model switches ensures that the filtering process remains efficient and reliable, even in the presence of challenging and abrupt transitions.

\begin{figure}[H]
    \centering
    \includegraphics[width=0.6\textwidth]{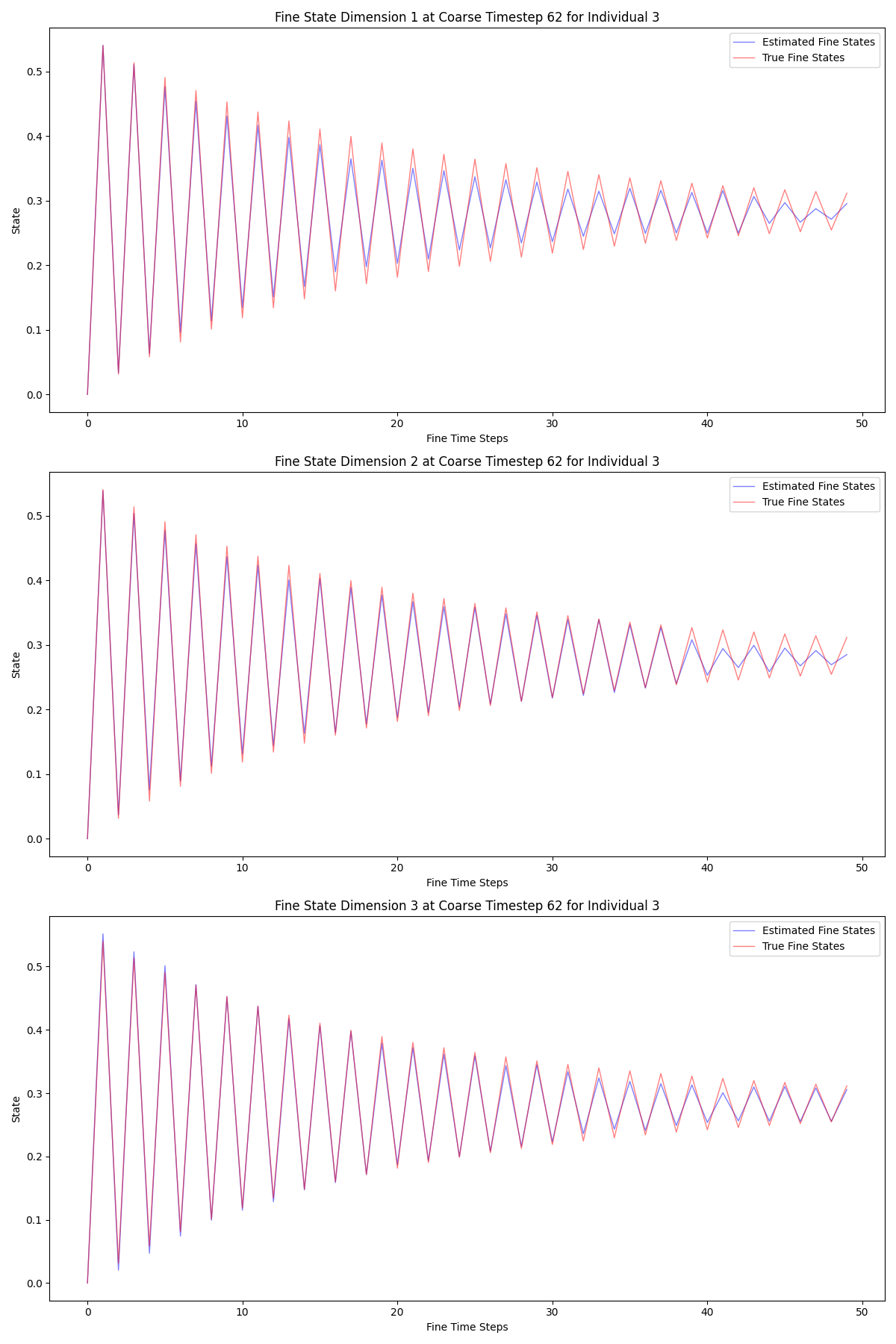}
    \caption{Comparison of true states and estimated states for Individual 3 at $t_{2} = 62$}. 
    \label{fig:fine_individual_3_timestep_62}
\end{figure}

The RMSE values for the $l=2$ time scale, summarized in Table \ref{tab:coarse_rmse}, further quantify the performance of the state estimation across the six individuals and three dimensions. These values reveal that the RMSE is relatively low for all individuals, ranging from approximately 0.111 to 0.157 across the three dimensions. Individuals $d=1$ through $d=3$ exhibit slightly higher RMSE values compared to Individuals $d=4$ through $d=6$, with the highest RMSE being observed for Individual $d=3$ in Coarse Dimension 1 at 0.157. Conversely, Individual $d=5$ has the lowest overall RMSE across all dimensions, with its Coarse Dimension 3 value reaching as low as 0.111. The differences in RMSE across individuals and dimensions suggest some variability in the estimation performance depending on the specific characteristics of the time series and the noise levels for each individual. Despite this variability, the relatively small range of RMSE values across all individuals indicates that the particle filter maintains a consistent level of accuracy in capturing the coarse states, regardless of the individual's specific time series dynamics. The robustness in performance, as reflected in these RMSE values, complements the visual analysis of the state trajectories and transitions between model indicators, reinforcing the overall reliability of the algorithm in handling multiscale state estimation. In addition to the results for $l=2$, the RMSE values for time scale $l=1$, presented in Tables \ref{tab:results_1_to_5}-\ref{tab:results_98_to_100} in the Appendix, indicate a consistent and accurate performance across the individuals. This consistency is further illustrated by Figure \ref{fig:fine_individual_3_timestep_62}. Given the number of figures, we have selected just one to highlight these findings, as the results are similar across all $t_{2}=1,...,100$.

\begin{table}[H]
    \centering
    \begin{tabular}{|c|c|c|c|}
        \hline
        \textbf{Individual} & \textbf{Dim 1} & \textbf{Dim 2} & \textbf{Dim 3} \\ 
        \hline
        Individual $d=1$ & 0.125 & 0.124 & 0.149 \\ 
        \hline
        Individual $d=2$ & 0.150 & 0.151 & 0.153 \\ 
        \hline
        Individual $d=3$ & 0.157 & 0.150 & 0.156 \\ 
        \hline
        Individual $d=4$ & 0.149 & 0.162 & 0.155 \\ 
        \hline
        Individual $d=5$ & 0.135 & 0.136 & 0.111 \\ 
        \hline
        Individual $d=6$ & 0.131 & 0.129 & 0.147 \\ 
        \hline
    \end{tabular}
    \caption{RMSE values for $l=2$ time scale across the three dimensions for each individual.}
    \label{tab:coarse_rmse}
\end{table}

\subsection{Simulation 2}
For the second set of simulations, the true dynamics for $l=2$ follow individual-specific switching patterns, where each individual $d$ alternates between two models, $m=0$ and $m=1$, at different points in time. For $d=1$, the dynamics follow model $m=1$ during the first third of the simulation (time steps $t = 1$ to $t = 33$), model $m=0$ in the second third (time steps $t = 34$ to $t = 66$), and return to model $m=1$ in the final third (time steps $t = 67$ to $t = 100$). Individual $d=2$ follows a different pattern, starting with model $m=0$ for the first third, switching to model $m=1$ for the second third, and then returning to model $m=0$ in the final third. Individual $d=3$ starts with model $m=0$, switches to model $m=1$ during the second third, and remains in model $m=1$ for the final third. Individual $d=4$ follows model $m=1$ for both the first and second thirds of the time series, before switching to model $m=0$ in the final third. Individual $d=5$ remains in model $m=0$ during the first two-thirds of the simulation, switching to model $m=1$ only in the final third. Lastly, Individual $d=6$ starts with model $m=1$, switches to model $m=0$ for the second third, and remains in model $m=0$ for the final third.

\subsubsection{Results and Discussion}
\begin{figure}[htbp]
    \centering
    \subfigure[Individual 1: State Estimate with $\Sigma_{\mathbf{w}_{1,t_2}^{[2]}} = 0.5 \, \mathbf{I}_{N_2}$]{
        \includegraphics[width=0.4\textwidth]{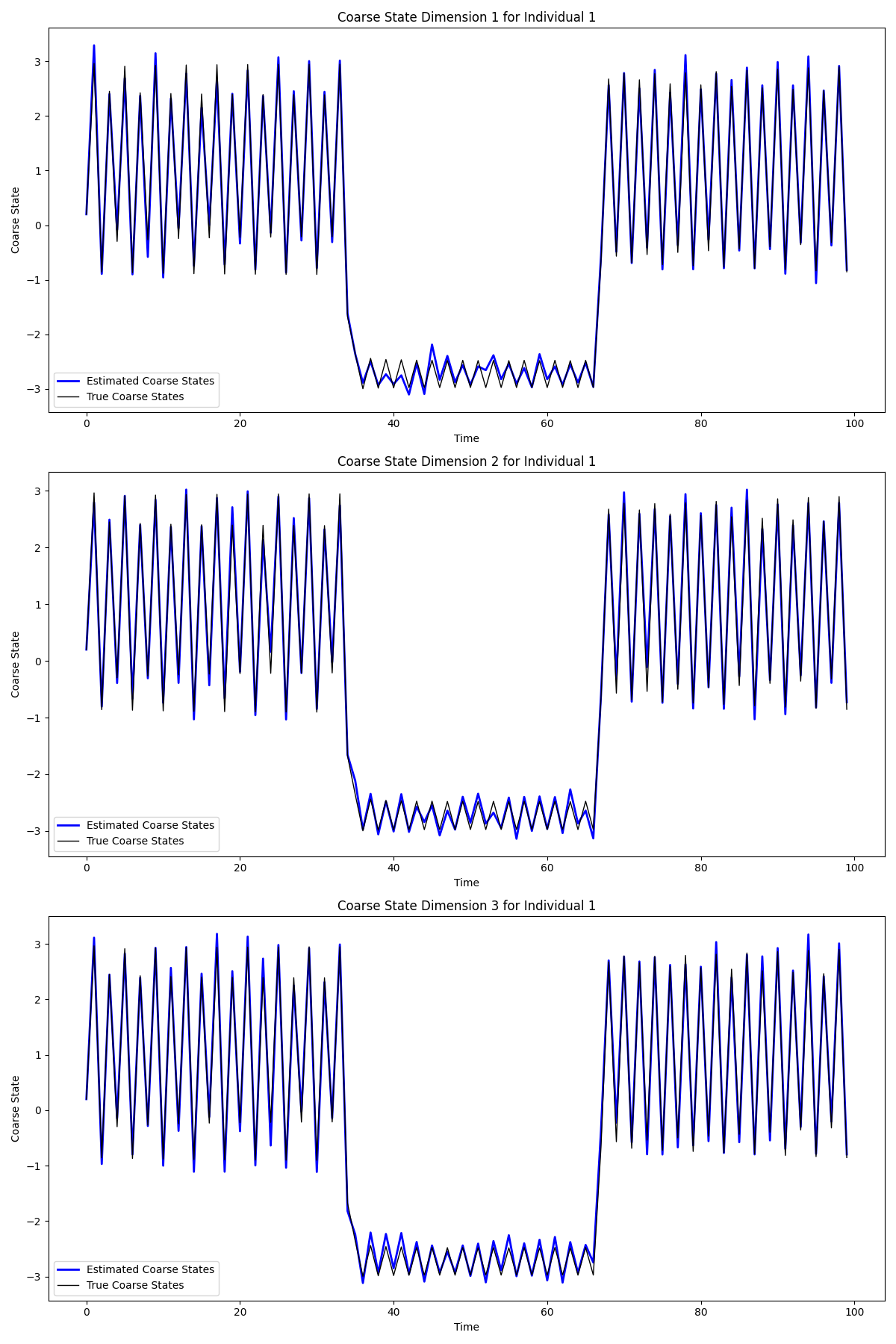}
        \label{fig:sim2_individual1_state}
    }
    \subfigure[Individual 1: True vs. Estimated Indicator with $\Sigma_{\mathbf{w}_{1,t_2}^{[2]}} = 0.5 \, \mathbf{I}_{N_2}$]{
        \includegraphics[width=0.8\textwidth]{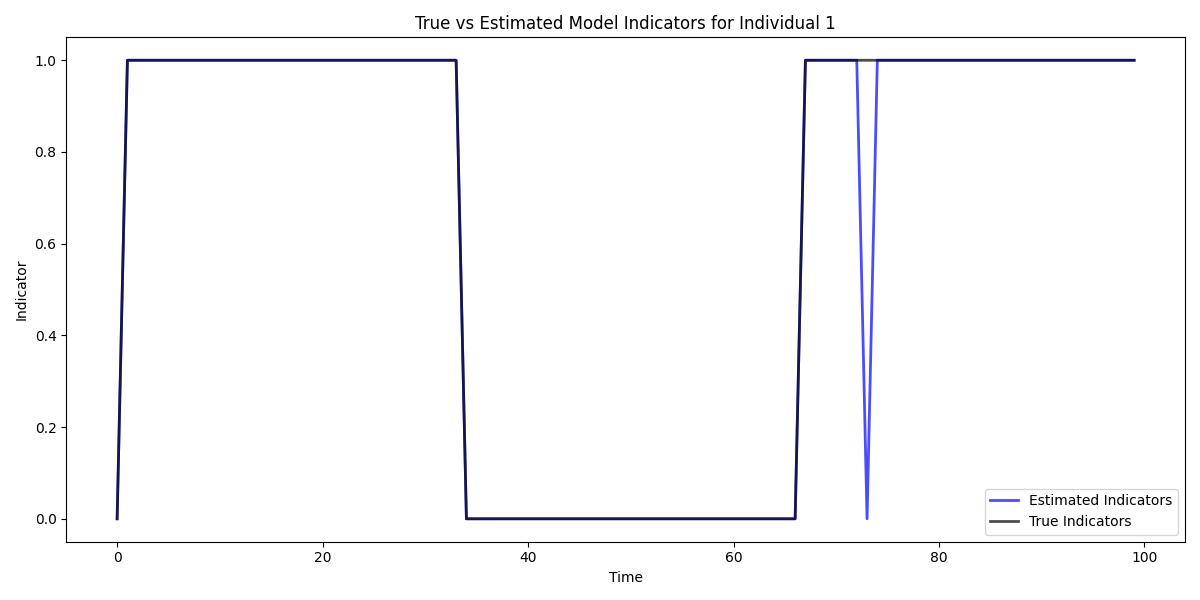}
        \label{fig:sim2_individual1_indicator}
    }
    \captionsetup{width=0.9\textwidth}
    \caption{Comparison of true states and estimated states for Individual 1 (Simulation 2), along with the estimated indicator. In the indicator plot, blue indicates the estimated model indicator while black represents the true model indicator.}
    \label{fig:sim2_true_vs_estimated_states_individual1}
\end{figure}

\begin{figure}[htbp]
    \centering
    \subfigure[Individual 2: State Estimate with $\Sigma_{\mathbf{w}_{2,t_2}^{[2]}} = 0.4 \, \mathbf{I}_{N_2}$]{
        \includegraphics[width=0.45\textwidth]{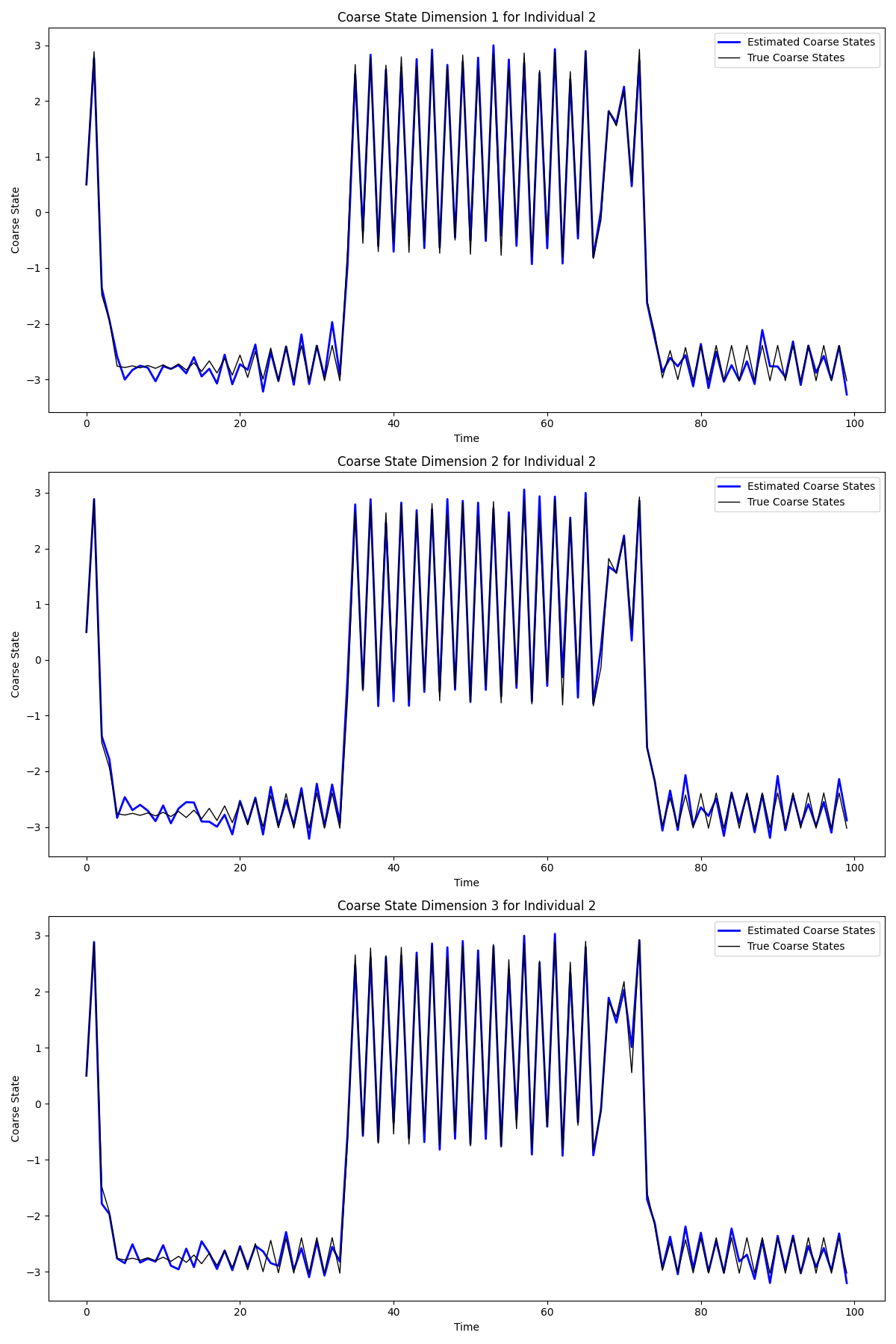}
        \label{fig:sim2_individual2_state}
    }
    \subfigure[Individual 2: True vs. Estimated Indicator with $\Sigma_{\mathbf{w}_{2,t_2}^{[2]}} = 0.4 \, \mathbf{I}_{N_2}$]{
        \includegraphics[width=0.8\textwidth]{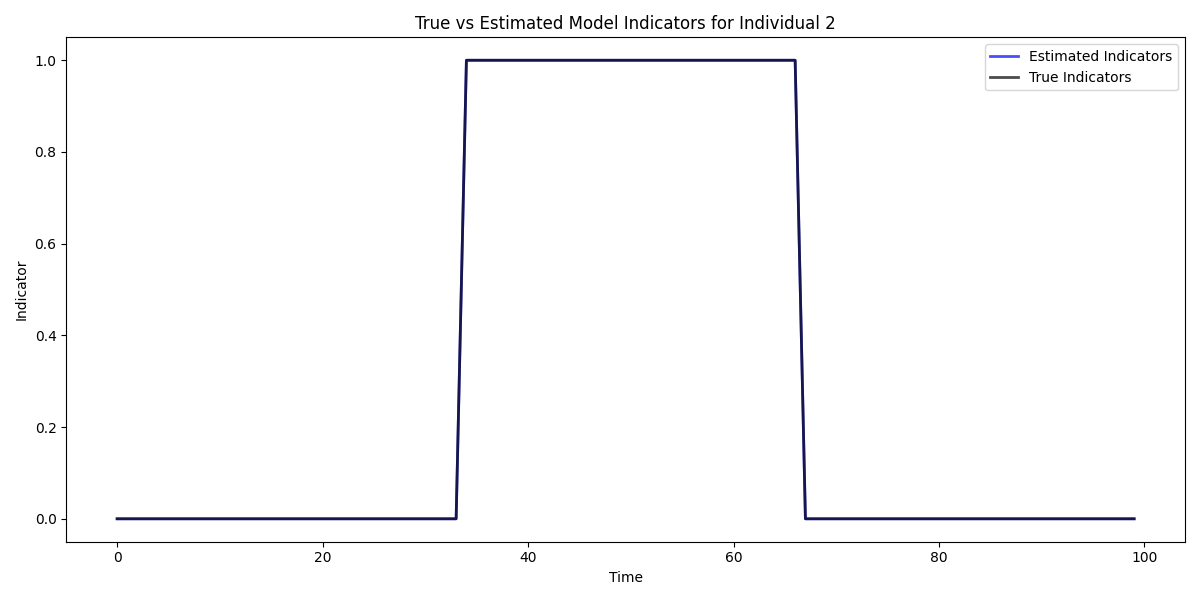}
        \label{fig:sim2_individual2_indicator}
    }
    \captionsetup{width=0.9\textwidth}
    \caption{Comparison of true states and estimated states for Individual 2 (Simulation 2), along with the estimated indicator. In the indicator plot, blue indicates the estimated model indicator while black represents the true model indicator.}
    \label{fig:sim2_true_vs_estimated_states_individual2}
\end{figure}

\begin{figure}[htbp]
    \centering
    \subfigure[Individual 3: State Estimate with $\Sigma_{\mathbf{w}_{3,t_2}^{[2]}} = 0.5 \, \mathbf{I}_{N_2}$]{
        \includegraphics[width=0.45\textwidth]{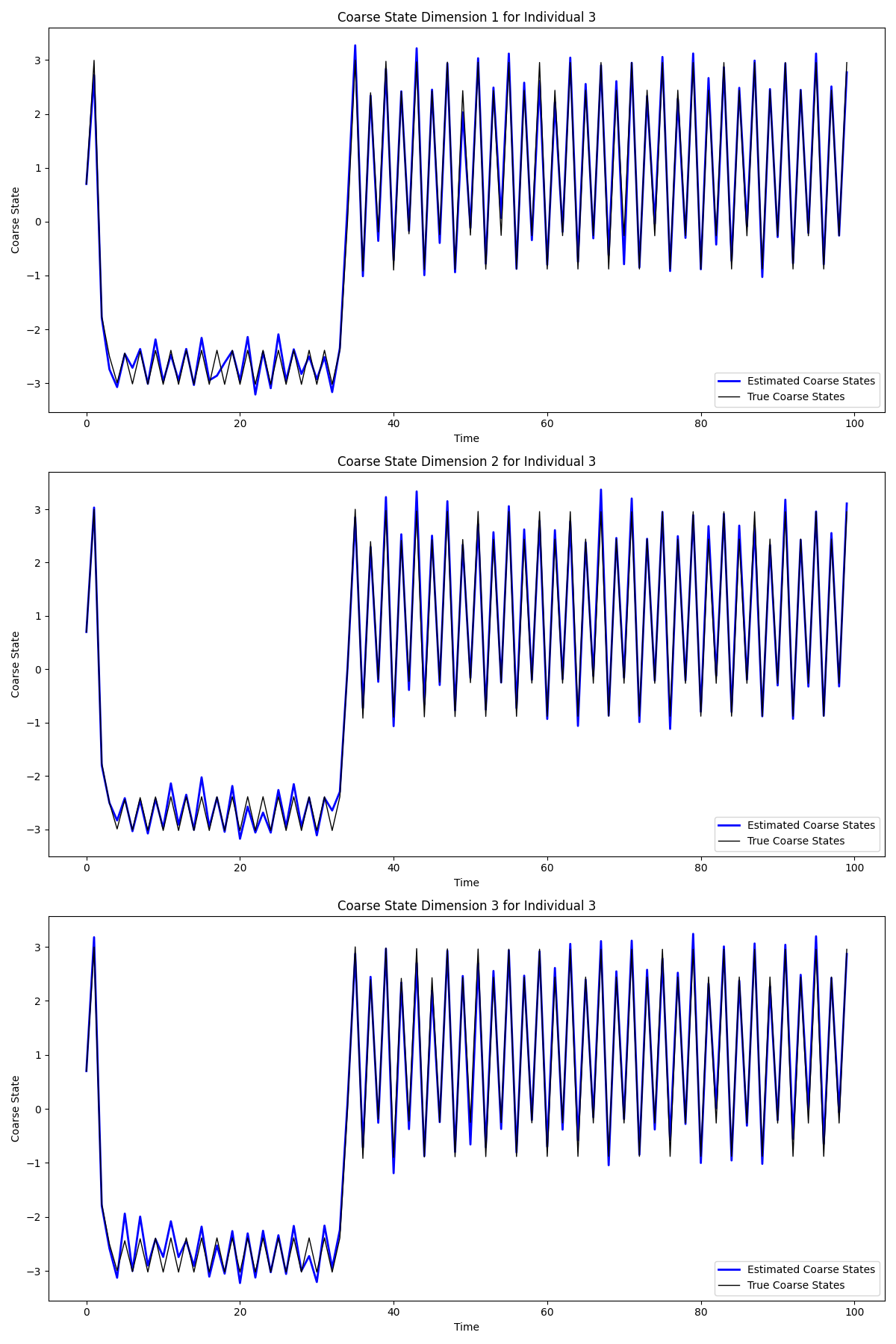}
        \label{fig:sim2_individual3_state}
    }
    \subfigure[Individual 3: True vs. Estimated Indicator with $\Sigma_{\mathbf{w}_{3,t_2}^{[2]}} = 0.5 \, \mathbf{I}_{N_2}$]{
        \includegraphics[width=0.8\textwidth]{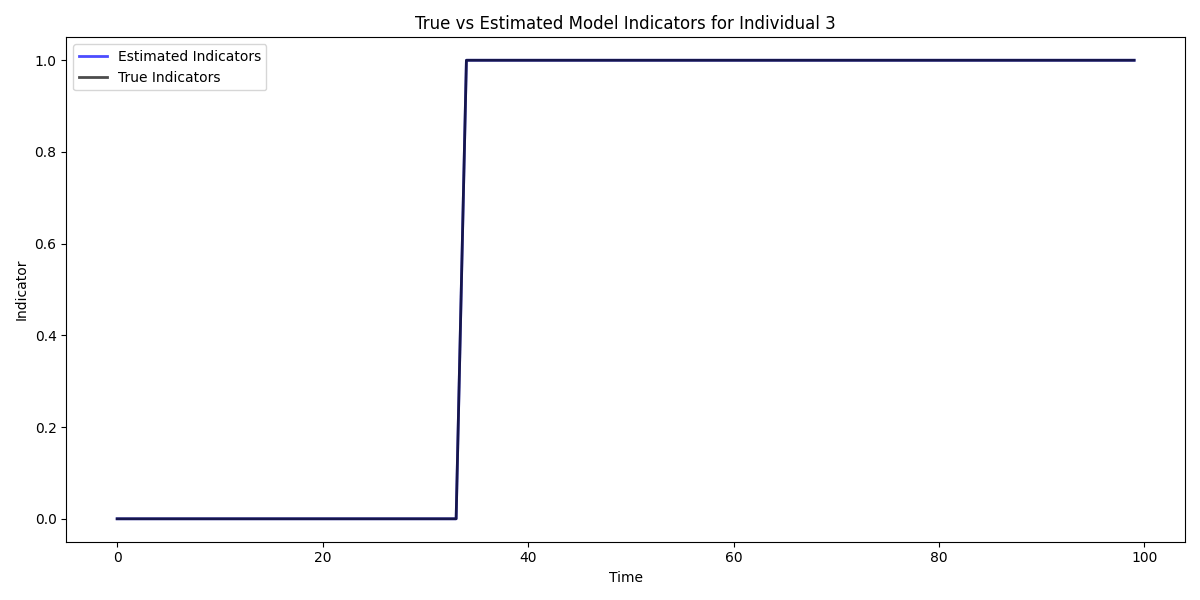}
        \label{fig:sim2_individual3_indicator}
    }
    \captionsetup{width=0.9\textwidth}
    \caption{Comparison of true states and estimated states for Individual 3 (Simulation 2), along with the estimated indicator. In the indicator plot, blue indicates the estimated model indicator while black represents the true model indicator.}
    \label{fig:sim2_true_vs_estimated_states_individual3}
\end{figure}

\begin{figure}[htbp]
    \centering
    \subfigure[Individual 4: State Estimate with $\Sigma_{\mathbf{w}_{4,t_2}^{[2]}} = 0.7 \, \mathbf{I}_{N_2}$]{
        \includegraphics[width=0.45\textwidth]{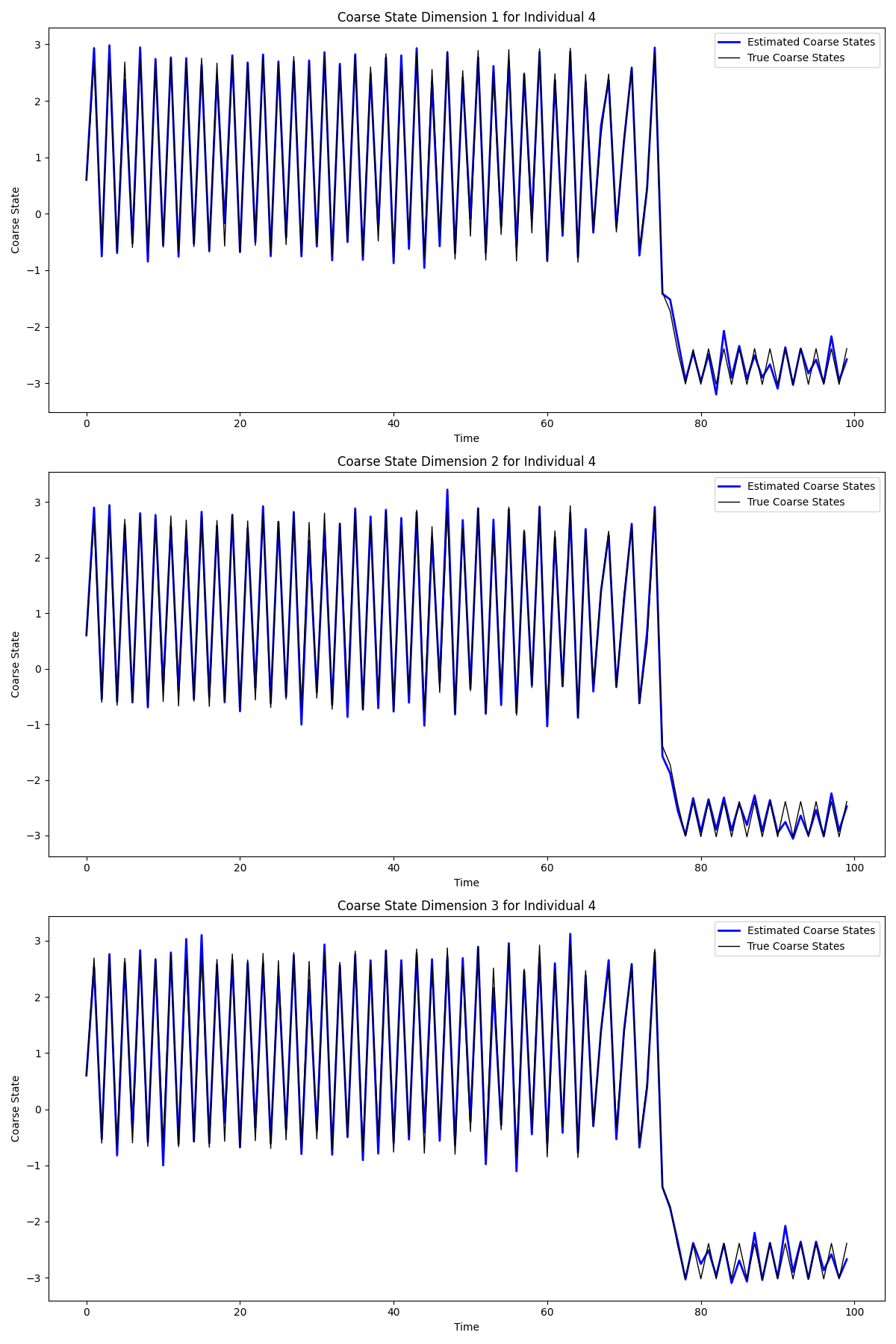}
        \label{fig:sim2_individual4_state}
    }
    \subfigure[Individual 4: True vs. Estimated Indicator with $\Sigma_{\mathbf{w}_{4,t_2}^{[2]}} = 0.7 \, \mathbf{I}_{N_2}$]{
        \includegraphics[width=0.8\textwidth]{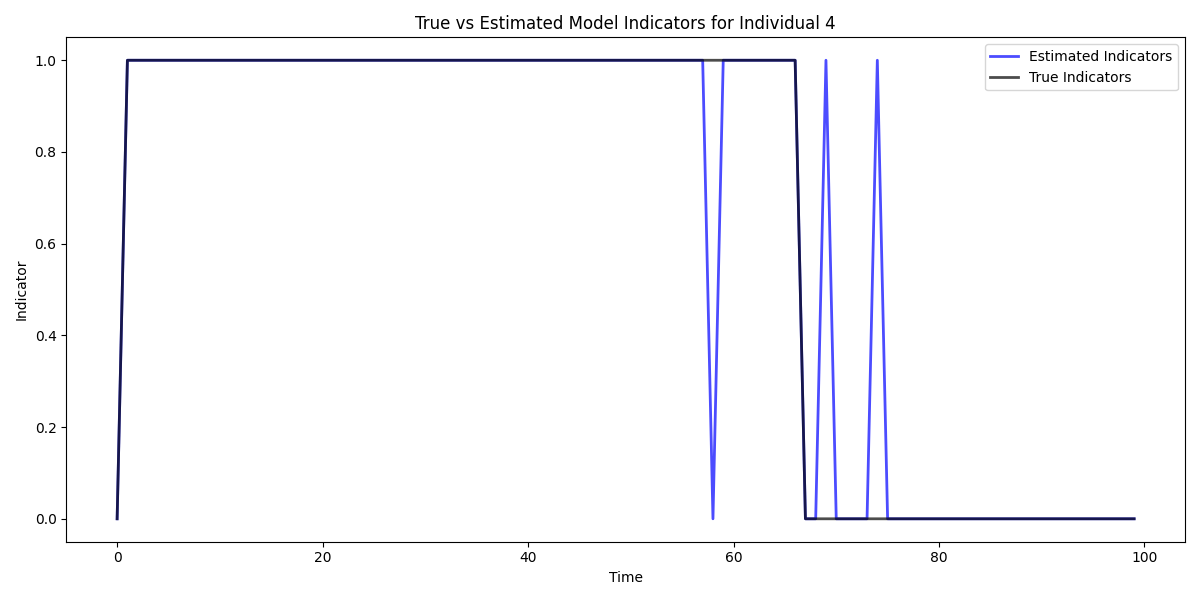}
        \label{fig:sim2_individual4_indicator}
    }
    \captionsetup{width=0.9\textwidth}
    \caption{Comparison of true states and estimated states for Individual 4 (Simulation 2), along with the estimated indicator. In the indicator plot, blue indicates the estimated model indicator while black represents the true model indicator.}
    \label{fig:sim2_true_vs_estimated_states_individual4}
\end{figure}

\begin{figure}[htbp]
    \centering
    \subfigure[Individual 5: State Estimate with $\Sigma_{\mathbf{w}_{5,t_2}^{[2]}} = 0.3 \, \mathbf{I}_{N_2}$]{
        \includegraphics[width=0.45\textwidth]{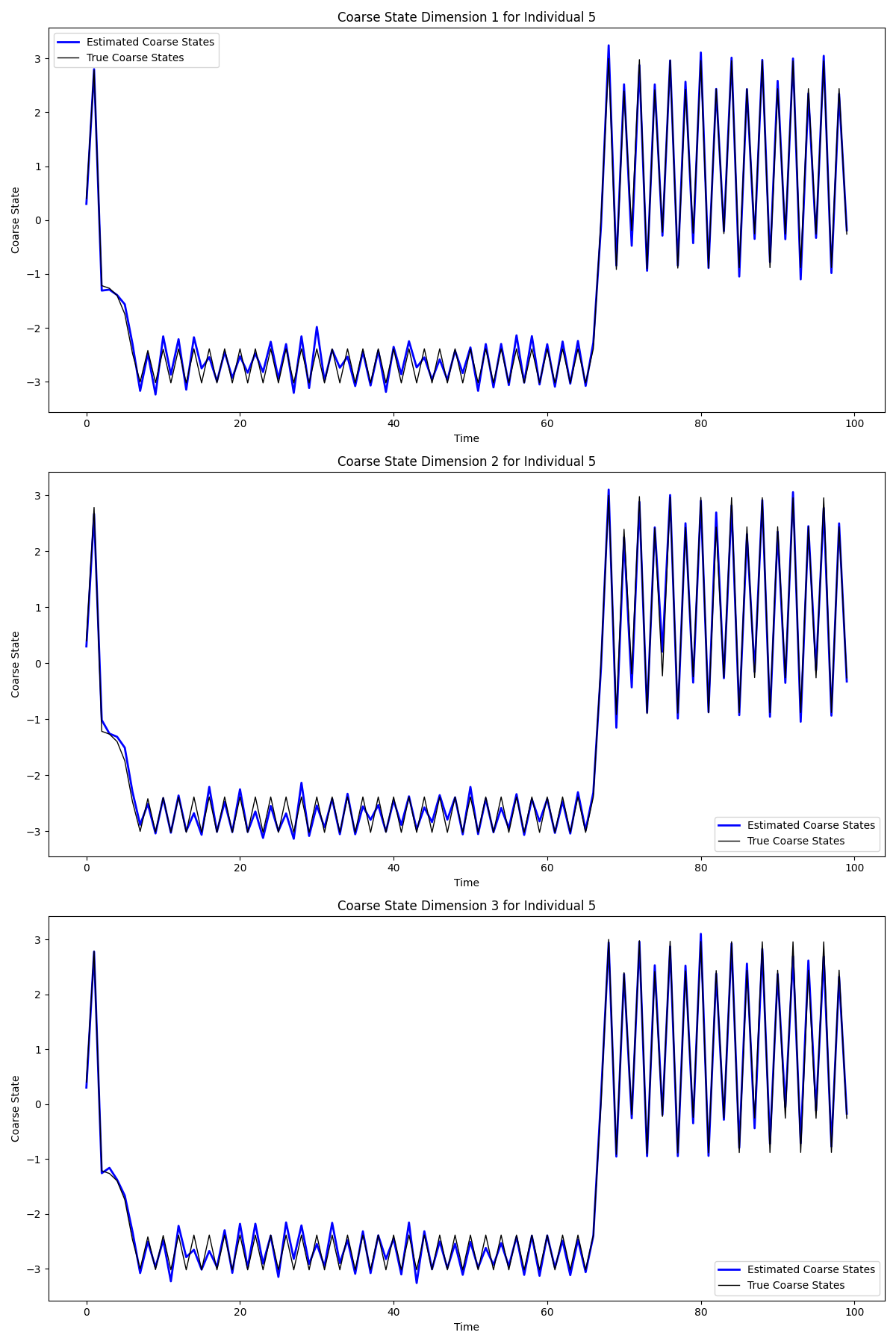}
        \label{fig:sim2_individual5_state}
    }
    \subfigure[Individual 5: True vs. Estimated Indicator with $\Sigma_{\mathbf{w}_{5,t_2}^{[2]}} = 0.3 \, \mathbf{I}_{N_2}$]{
        \includegraphics[width=0.8\textwidth]{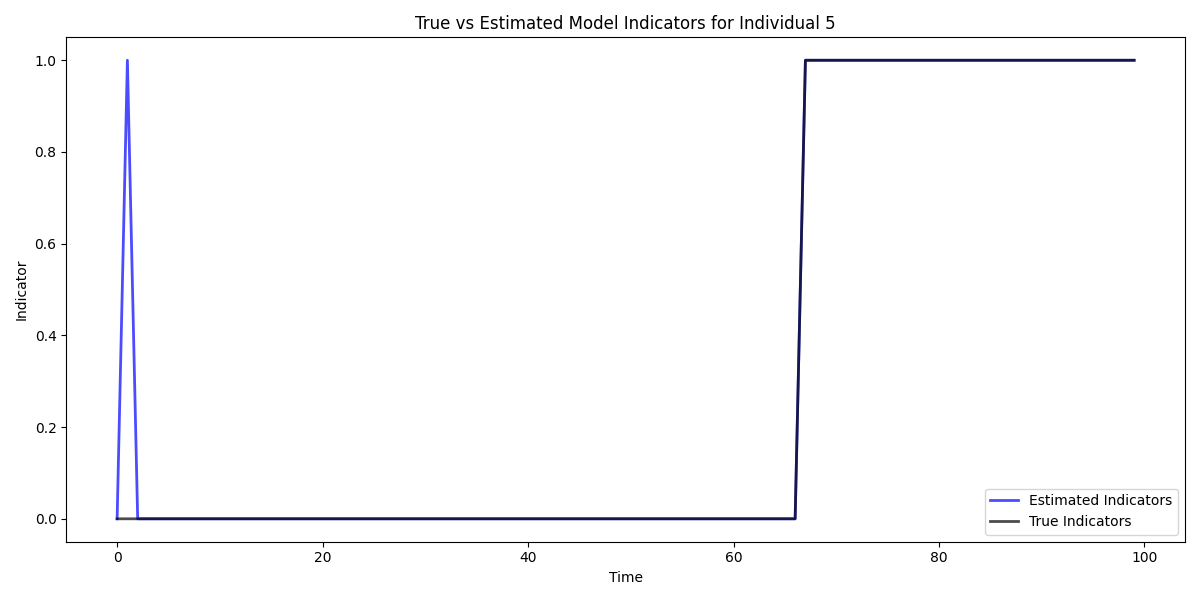}
        \label{fig:sim2_individual5_indicator}
    }
    \captionsetup{width=0.9\textwidth}
    \caption{Comparison of true states and estimated states for Individual 5 (Simulation 2), along with the estimated indicator. In the indicator plot, blue indicates the estimated model indicator while black represents the true model indicator.}
    \label{fig:sim2_true_vs_estimated_states_individual5}
\end{figure}

\begin{figure}[htbp]
    \centering
    \subfigure[Individual 6: State Estimate with $\Sigma_{\mathbf{w}_{6,t_2}^{[2]}} = 0.4 \, \mathbf{I}_{N_2}$]{
        \includegraphics[width=0.45\textwidth]{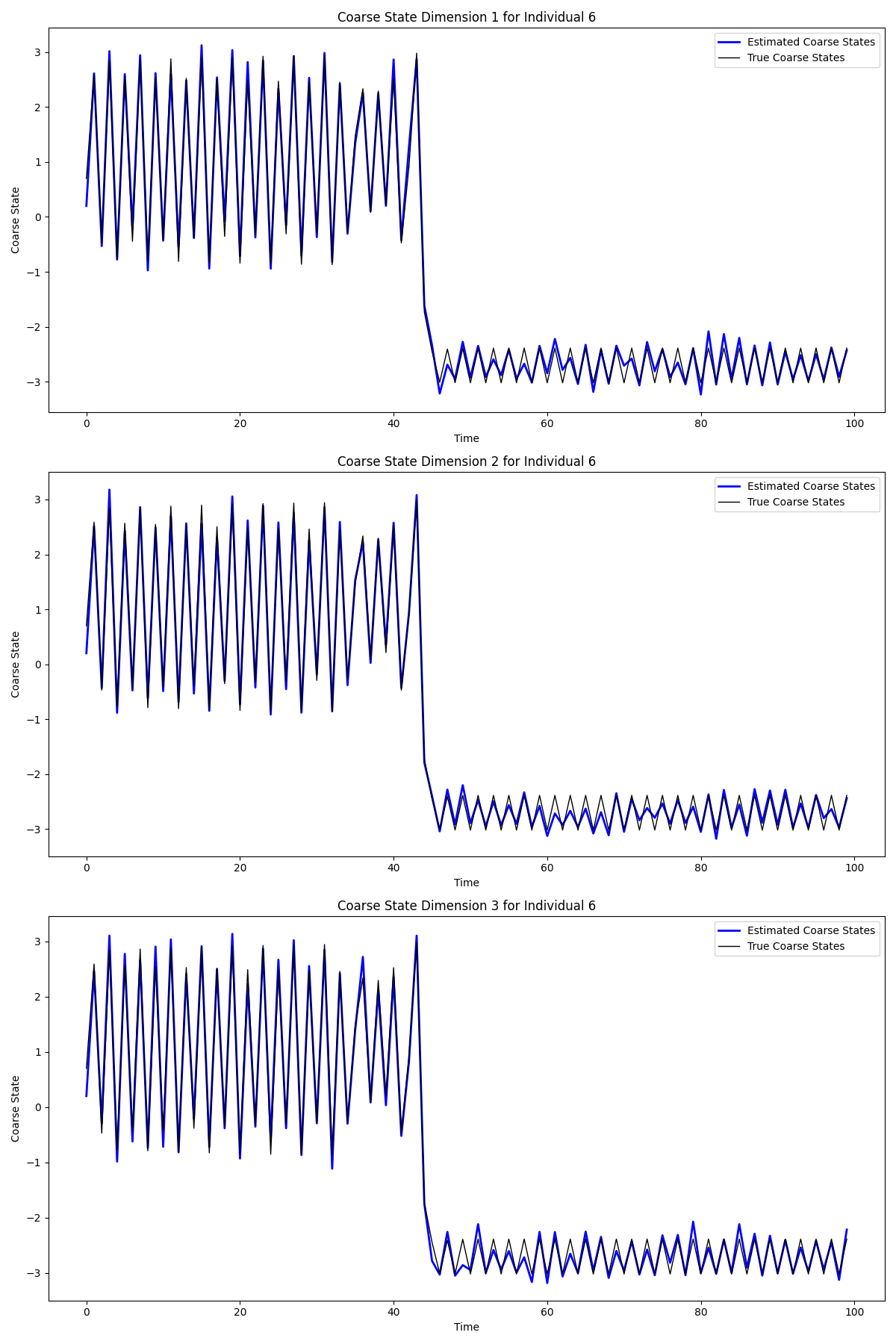}
        \label{fig:sim2_individual6_state}
    }
    \subfigure[Individual 6: True vs. Estimated Indicator with $\Sigma_{\mathbf{w}_{6,t_2}^{[2]}} = 0.4 \, \mathbf{I}_{N_2}$]{
        \includegraphics[width=0.8\textwidth]{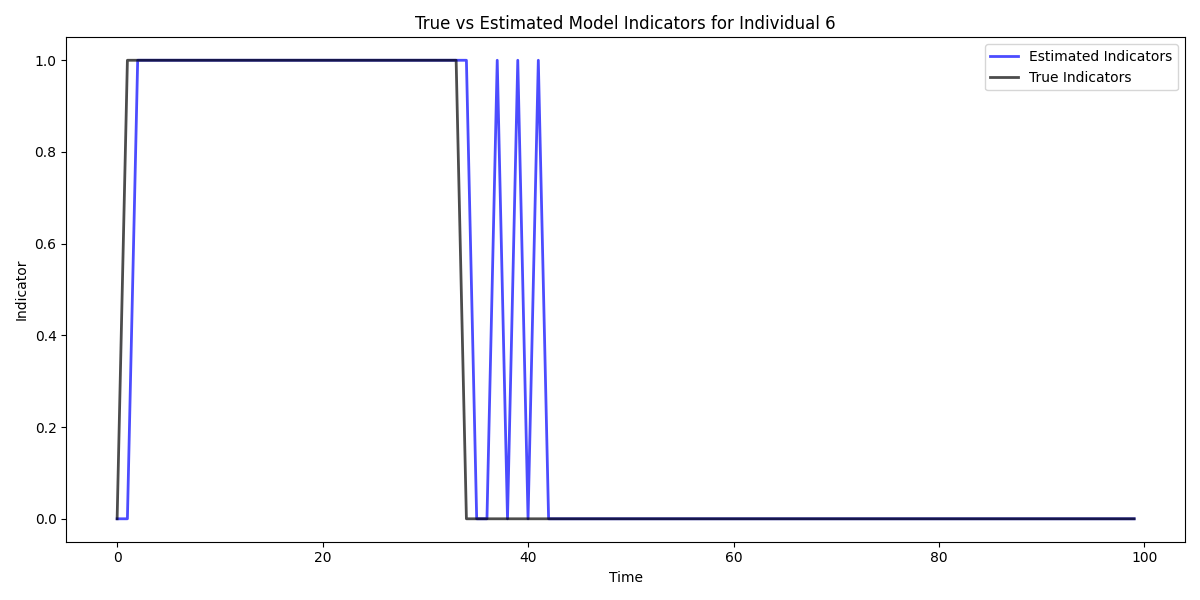}
        \label{fig:sim2_individual6_indicator}
    }
    \captionsetup{width=0.9\textwidth}
    \caption{Comparison of true states and estimated states for Individual 6 (Simulation 2), along with the estimated indicator. In the indicator plot, blue indicates the estimated model indicator while black represents the true model indicator.}
    \label{fig:sim2_true_vs_estimated_states_individual6}
\end{figure}

\begin{figure}[H]
    \centering
    \includegraphics[width=0.6\textwidth]{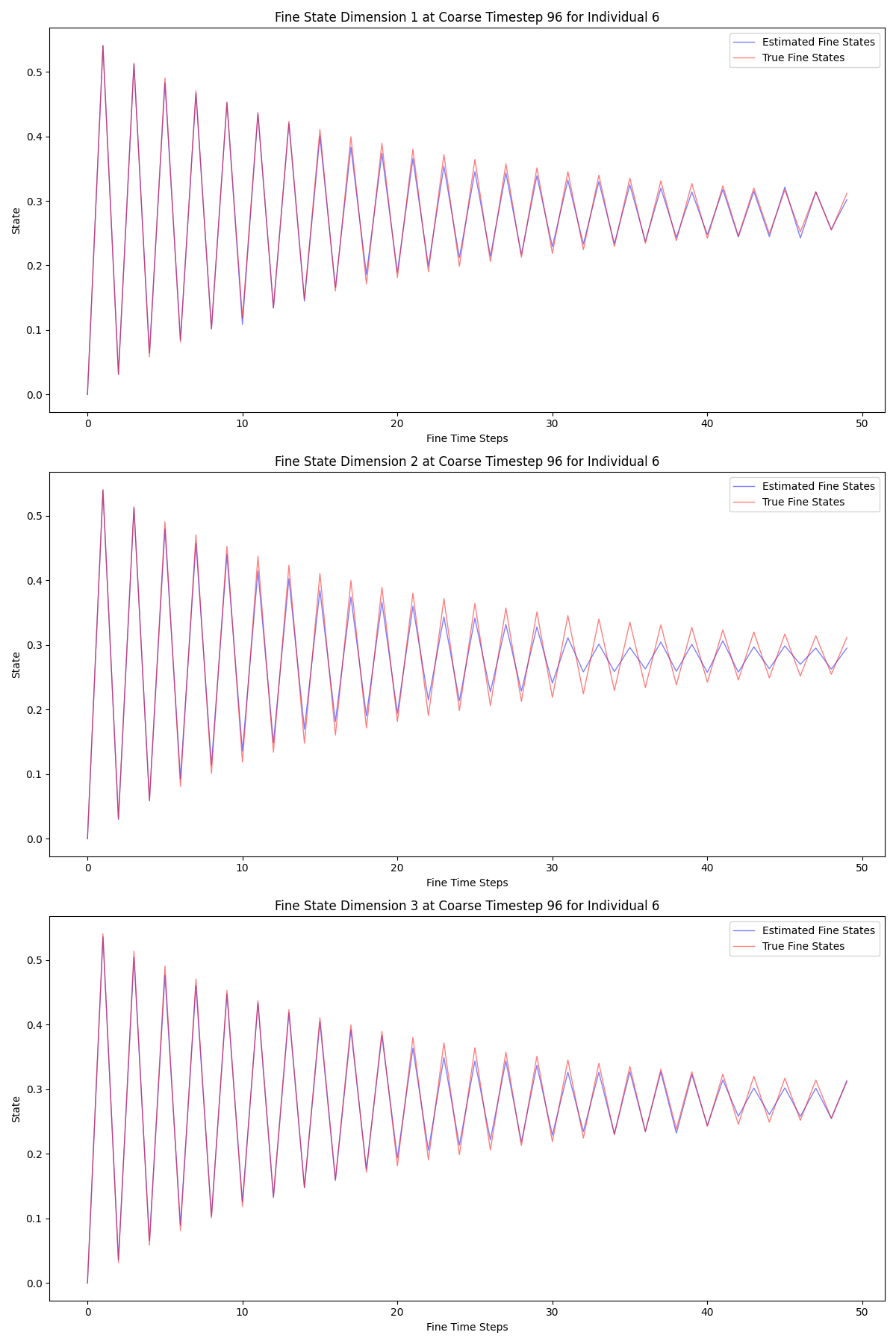}
    \caption{Comparison of true states and estimated states for Individual 3 at $t_{2} = 96$ (Simulation 2)}. 
    \label{fig:fine_individual_6_timestep_96}
\end{figure}

In this section, we evaluate the performance of the proposed model for six different individuals by comparing the true states and the estimated states for each individual, as well as the accuracy of the estimated model indicators. Figures \ref{fig:sim2_individual1_state} to \ref{fig:sim2_individual6_state} display the true and estimated states for all dimensions of individuals $d=1$ to $d=6$. The figures reveal that, overall, the particle filter accurately tracks the true coarse-scale states, with only minor discrepancies occurring in some individuals during transitions between dynamic regimes. Because the fine time scale state estimates and RMSE results for Simulation 2 are very similar to those for Simulation 1, we have omitted the full set of results and present only one figure to demonstrate the model's performance. This representative figure (Figure \ref{fig:fine_individual_6_timestep_96}) highlights the consistency in the model's ability to estimate fine time scale states accurately across different simulations.

For $d=1$ (Figure \ref{fig:sim2_individual1_state}), the estimated states closely match the true states, with some minor deviations visible during the transitions between models around $t=33$ and $t=66$, but the filter quickly recovers and follows the true dynamics well. Next, $d=2$ (Figure \ref{fig:sim2_individual2_state}) shows excellent alignment between the true and estimated states. The filter manages to capture the rapid transitions between models at $t=33$ and $t=66$ accurately, reflecting the capability of the model to adapt to abrupt changes in dynamics. In the case of $d=3$ (Figure \ref{fig:sim2_individual3_state}), the estimated states again show close alignment with the true states, with the filter accurately capturing both the oscillatory dynamics and the transitions between models. individuals $d=4$ (Figure \ref{fig:sim2_individual4_state}), $d=5$ (Figure \ref{fig:sim2_individual5_state}), and $d=6$ (Figure \ref{fig:sim2_individual6_state}) all exhibit similar behavior, with the particle filter accurately estimating the coarse states. Although some minor errors are observed during rapid transitions in $d=4$ and $d=6$, these errors are quickly corrected by the model, resulting in good overall performance. Table \ref{tab:coarse_rmse_sim2} presents the RMSE values for each individual across the three coarse-scale dimensions. The values indicate that, overall, the particle filter provides accurate state estimates across all dimensions and individuals, with RMSE values remaining relatively low and consistent. Individuals $d=1$, $d=4$, and $d=5$ show the lowest RMSE values, suggesting that the filter performed particularly well for these individuals. In contrast, $d=3$ and $d=6$ have slightly higher RMSE values, which may reflect the challenges the filter faced in capturing rapid transitions in their dynamic regimes. Despite these small variations, the results confirm the model's robustness in estimating the coarse-scale states for all individuals.

Figures \ref{fig:sim2_individual1_indicator} to \ref{fig:sim2_individual6_indicator} show the true and estimated model indicators for individuals $d=1$ to $d=6$. These figures highlight the model's ability to correctly infer the underlying dynamic model that each individual follows over time. For $d=1$ (Figure \ref{fig:sim2_individual1_indicator}), the estimated indicators align closely with the true indicators, except for a brief discrepancy around $t=70$, where the filter temporarily assigns the wrong model before correcting itself. Despite this, the filter demonstrates high accuracy overall. Individuals $d=2$ (Figure \ref{fig:sim2_individual2_indicator}) and $d=3$ show strong alignment between the estimated and true indicators, demonstrating the model's ability to handle rapid transitions between models effectively. No significant deviations are observed. For $d=4$ (Figure \ref{fig:sim2_individual4_indicator}), a few errors are observed around $t_{2}=60$ and shortly after $t_{2}=70$. However, these errors are temporary, and the model eventually converges to the correct indicator. As well, the transition between the models is accurately estimated. For $d=5$ (Figure \ref{fig:sim2_individual5_indicator}), the estimated indicators show close alignment with the true indicators, with only brief discrepancy in the beginning of the time series that is quickly corrected. However, for individual $d=6$ (Figure \ref{fig:sim2_individual6_indicator}), the model encounters more significant difficulties in detecting the transitions. Around $t_{2} = 30$, the estimated indicators rapidly oscillate between models before settling into the correct indicator after several time steps. This suggests that while the model is capable of eventually identifying the correct dynamics, it struggles with rapid changes in the underlying model, leading to transient errors in the model indicators during periods of transition. Despite these issues, the model ultimately converges to the correct state. 

The results indicate that our proposed learning approach performs well in estimating both the fine and coarse-scale states and the model indicators for all individuals. While some minor discrepancies are observed during periods of rapid transition between models, the filter is generally able to recover and follow the true dynamics accurately. The model is particularly effective in estimating the correct model indicators, with only brief errors occurring in a few individuals.

\begin{table}[H]
    \centering
    \begin{tabular}{|c|c|c|c|}
        \hline
        \textbf{Individual} & \textbf{Dim 1} & \textbf{Dim 2} & \textbf{Dim 3} \\ 
        \hline
        Individual $d=1$ & 0.134 & 0.133 & 0.142 \\ 
        \hline
        Individual $d=2$ & 0.153 & 0.157 & 0.146 \\ 
        \hline
        Individual $d=3$ & 0.165 & 0.146 & 0.170 \\ 
        \hline
        Individual $d=4$ & 0.139 & 0.143 & 0.159 \\ 
        \hline
        Individual $d=5$ & 0.138 & 0.129 & 0.126 \\ 
        \hline
        Individual $d=6$ & 0.152 & 0.143 & 0.164 \\ 
        \hline
    \end{tabular}
    \caption{RMSE values for $l=2$ time scale across the three dimensions for each individual.}
    \label{tab:coarse_rmse_sim2}
\end{table}

\clearpage
\subsection{Conclusion}
The multiscale state-space modeling framework developed in this work offers a novel and versatile approach for capturing the dynamics of complex systems. A key innovation of this framework is its ability to model nested nonlinear dynamics, where processes at different levels interact and influence each other. By effectively modeling interactions between different scales, the framework offers a powerful tool for understanding systems with intricate, layered dynamics, where both gradual evolutions and abrupt transitions are critical.

The results demonstrate that our Bayesian learning model, based on this multiscale framework, successfully tracks states and identifies the correct model indicators across all individuals with little variation. The accuracy observed in both state estimates and model indicators highlights the robustness of the framework's switching mechanism, which can handle diverse dynamic regimes. This ability to learn transitions in dynamics is essential for managing complex systems, as these shifts can lead to significant and often irreversible changes in behavior.

Moving forward, a key direction for future work is the integration of nonparametric approaches, such as Dirichlet and Gaussian process priors, to allow for a more flexible model selection process. By utilizing nonparametric methods, we can remove the assumption of a fixed number of known models and enable the model to adaptively discover the appropriate number of dynamic regimes. This could further enhance our framework's ability to capture complex, evolving dynamics, particularly in systems where the number of underlying states or models is unknown or variable. 
\clearpage
\bibliographystyle{unsrt}

\newpage
\section{Appendix A: Derivations for Conditional Distributions}
We begin with the full joint posterior for the model, which is written as:
\begin{align}
    \prod_{d=1}^{D} \prod_{m=1}^{M} \prod_{l=1}^{L} \prod_{t_{l}=1}^{T_{l}} p\left( \mathbf{x}_{d,t_{l}}^{[l]}, \mathbf{s}_{d,t_{L}}^{[L]}, \pi_{d,t_{L}}^{L,(m)} \vert \mathbf{y}_{d,t_{l}}^{[l]}, \alpha_{m}, \Sigma_{\mathbf{w}_{d,t_{l}}^{[l]}}, \Sigma_{\mathbf{v}_{d,t_{l}}^{[l]}} \right)
\end{align}
Here, the matrices $\Sigma_{\mathbf{w}_{d,t_{l}}^{[l]}}$ and $ \Sigma_{\mathbf{v}_{d,t_{l}}^{[l]}}$ are known and do not have associated probability density functions. The term \( s_{d,t_{L}}^{[L]} \) represents the indicator for the model, and \( \pi_{d,t_{L}}^{L,(m)} \) represents the probabilities associated with selecting the indicators.

\subsection{Conditional Posterior of the Model Probabilities  \( \boldsymbol{\pi}_{d,t_{L}}^{[L]} \)}
The full conditional for the model probabilities \( \boldsymbol{\pi}_{d,t_{L}}^{[L]} \) can be derived by considering the joint posterior and applying Bayes' theorem. We take into account the number of times each indicator \( \mathbf{s}_{d,t_{L}}^{[L]} \) was selected. The posterior depends on both the prior for \( \boldsymbol{\pi}_{d,t_{L}}^{[L]} \) (a Dirichlet distribution) and the categorical distribution for \( \mathbf{s}_{d,t_{L}}^{[L]} \), which is conditional on the previous indicators \( \mathcal{S}_{d,t_{L}-1}^{[L]} \) and the probabilities \( \boldsymbol{\pi}_{d,t_{L}}^{[L]} \). We write the full conditional posterior for \( \boldsymbol{\pi}_{d,t_{L}}^{[L]} \) as follows:
\begin{equation}
p(\boldsymbol{\pi}_{d,t_{L}}^{[L]} \vert \mathbf{s}_{d,t_{L}}^{[L]}, \boldsymbol{\alpha}) \propto p(\mathbf{s}_{d,t_{L}}^{[L]} \vert \mathcal{S}_{d,1:t_{L}-1}^{[L]}, \boldsymbol{\pi}_{d,t_{L}}^{[L]}) p(\boldsymbol{\pi}_{d,t_{L}}^{[L]} \vert \boldsymbol{\alpha})
\end{equation}
The indicators \( \mathbf{s}_{d,t_{L}}^{[L]} \) are drawn from a categorical distribution. The probability of selecting a specific model at time \( t_{L} \) depends on the set of past indicators \( \mathcal{S}_{d,t_{L}-1}^{[L]} \) for time $t_{L}=1,...,t_{L}-1$ and the probabilities \( \boldsymbol{\pi}_{d,t_{L}}^{[L]} \). The conditional distribution for \( \mathbf{s}_{d,t_{L}}^{[L]} \) is written as:
\begin{equation}
p(\mathbf{s}_{d,t_{L}}^{[L]} \vert \mathcal{S}_{d,t_{L}-1}^{[L]}, \boldsymbol{\pi}_{d,t_{L}}^{[L]}) = \prod_{t_{L}=1}^{T_{L}} \text{Cat}\left( \mathbf{s}_{d,t_{L}}^{[L]} \vert \mathcal{S}_{d,t_{L}-1}^{[L]}, \boldsymbol{\pi}_{d,t_{L}}^{[L]} \right)
\end{equation}
Here, \( \mathbf{s}_{d,t_{L}}^{[L]} \) follows a categorical distribution where the probabilities for selecting each model at time \( t_{L} \) are given by the vector \( \boldsymbol{\pi}_{d,t_{L}}^{[L]} \). The prior on the model probabilities \( \boldsymbol{\pi}_{d,t_{L}}^{[L]} \) is a Dirichlet distribution with parameters \( \boldsymbol{\alpha} = (\alpha_1, \alpha_2, \dots, \alpha_M) \), encoding the prior belief about how likely each model is to be selected:
\begin{equation}
p(\boldsymbol{\pi}_{d,t_{L}}^{[L]} \vert \boldsymbol{\alpha}) = \text{Dir}(\alpha_{1}, \alpha_{2}, \dots, \alpha_{M})
\end{equation}
To derive the full conditional for \( \boldsymbol{\pi}_{d,t_{L}}^{[L]} \), we apply Bayes' theorem. By multiplying the probability of the indicators by the prior distribution of \( \boldsymbol{\pi}_{d,t_{L}}^{[L]} \), we have:
\begin{equation}
p(\boldsymbol{\pi}_{d,t_{L}}^{[L]} \vert \mathbf{s}_{d,t_{L}}^{[L]}, \boldsymbol{\alpha}) \propto \left( \prod_{t_{L}=1}^{T_{L}} \text{Cat}\left( \mathbf{s}_{d,t_{L}}^{[L]} \vert \mathcal{S}_{d,t_{L}-1}^{[L]}, \boldsymbol{\pi}_{d,t_{L}}^{[L]} \right) \right)\text{Dir}(\alpha_{1}, \alpha_{2}, \dots, \alpha_{M})
\end{equation}
Because the Dirichlet distribution is conjugate to the categorical distribution, the resulting posterior for \( \boldsymbol{\pi}_{d,t_{L}}^{[L]} \) is also Dirichlet. The updated parameters of this posterior are the sum of the prior parameters \( \alpha_m \) and the counts \( c_{d,1:t_{L}-1}^{[L],(m)} \), where \( c_{d,1:t_{L}-1}^{[L],(m)} \) is the number of times the indicator \( \mathbf{s}_{d,t_{L}}^{[L]} = m \) has been observed up to time \( t_{L}-1 \).
Thus, the full conditional is:
\begin{equation}
\boldsymbol{\pi}_{d,t_{L}}^{[L]} \vert \mathcal{S}_{d,t_{L}-1}^{[L]}, \mathbf{s}_{d,t_{L}}^{[L]} \sim \text{Dir}\left( \alpha_{1} + c_{d,1:t_{L}-1}^{[L],(1)}, \alpha_{2} + c_{d,1:t_{L}-1}^{[L],(2)}, \dots, \alpha_{M} + c_{d,1:t_{L}-1}^{[L],(M)} \right)
\end{equation}

\subsection{Conditional Posterior of the Model Indicators \( s_{d,t_{L}}^{[L]} \)}
Next, we derive the conditional posterior of the indicator \( s_{d,t_{L}}^{[L]} \):
\begin{align}
    p(s_{d,t_{L}}^{[L]} \vert \mathcal{S}_{d,t_{L}-1}^{[L]}, \mathbf{x}_{d,t_{L}}^{[L]}, \mathbf{y}_{d,t_{L}}^{[L]}, \pi_{d,t_{L}}^{L,(m)}),
\end{align}
Recall, the joint posterior for all unknowns, including states, indicators, and probabilities, is given by:
\begin{align}
    \prod_{d=1}^{D} \prod_{m=1}^{M} \prod_{l=1}^{L} \prod_{t_{l}=1}^{T_{l}} p\left( \mathbf{x}_{d,t_{l}}^{[l]}, \mathbf{s}_{d,t_{L}}^{[L]}, \pi_{d,t_{L}}^{L,(m)} \vert \mathbf{y}_{d,t_{l}}^{[l]}, \alpha_{m}, \Sigma_{\mathbf{w}_{d,t_{l}}^{[l]}}, \Sigma_{\mathbf{v}_{d,t_{l}}^{[l]}} \right)
\end{align}
where for brevity we have omitted the measurement and process noise covariance matrices as they are assumed to be known. Since the indicators \( \mathbf{s}_{d,t_{L}}^{[L]} \) only affect the coarse time scale \( L \), we can factor the joint posterior recursively by time steps \( T_L \). For each time step \( t_L \), and using the Markov assumption to factor recursively, the joint posterior can be written as:
\begin{align}
    &p(\mathbf{x}_{d,1:T_L}^{[L]}, \left\{ \mathbf{x}^{[l']}_{d,1:T_{l'}} \right\}_{l'=1}^{L-1}, \mathbf{s}_{d,1:T_L}^{[L]}, \pi_{d,1:T_L}^{[L]} \vert \mathbf{y}_{d,t_{L}}^{[L]}, \boldsymbol{\alpha}) \\
    &= p(\mathbf{x}_{d,t_L}^{[L]},  \left\{ \mathbf{x}^{[l']}_{d,1:T_{l'}} \right\}_{l'=1}^{L-1}, \mathbf{s}_{d,t_L}^{[L]}, \pi_{d,t_L}^{[L]} \vert \mathbf{y}_{d,1:t_L}^{[L]}, \mathbf{x}_{1:t_{L}-1}^{[L]}, \mathcal{S}_{d,t_{L}-1}^{[L]}, \pi_{d,1:t_{L}-1}^{[L]}, \boldsymbol{\alpha})\\
    &p(\mathbf{x}_{d,1:t_{L}-1}^{[L]}, \mathcal{S}_{d,t_{L}-1}^{[L]}, \pi_{d,1:t_L-1}^{[L]} \vert \mathbf{y}_{d,1:t_L-1}^{[L]})
\end{align}
In practice, during the particle filter update step, the joint posterior up to the previous time step \( t_{L-1} \) is taken into account.
This means that the term 
\[
p(\mathbf{x}_{d,1:t_{L}-1}^{[L]}, \mathcal{S}_{d,t_{L}-1}^{[L]}, \pi_{d,1:t_L-1}^{[L]} \mid \mathbf{y}_{d,1:t_L-1}^{[L]})
\]
represents the contribution of the previous time steps and is recursively updated using particle filtering. As new observations come in, the particle filter updates the posterior at the current time step \( t_L \) based on the filtered states and indicators from the previous time step. Using Bayes' theorem, the conditional posterior can be factored as follows:
\begin{align}
    p( s_{d,t_{L}}^{[L]} \mid \mathcal{S}_{d,t_{L}-1}^{[L]}, \mathbf{x}_{d,t_{L}}^{[L]}, 
    \mathbf{y}_{d,t_{L}}^{[L]}, \pi_{d,t_{L}}^{L,(m)} ) &\propto p( \mathbf{y}_{d,t_{L}}^{[L]} \mid \mathbf{x}_{d,t_{L}}^{[L]}) \nonumber \\
    &p( \mathbf{x}_{d,t_{L}}^{[L]} \mid s_{d,t_{L}}^{[L]}, \mathbf{x}_{d,t_{L}-1}^{[L]}, 
    \left\{ \mathbf{x}^{[l']}_{d,1:T_{l'}} \right\}_{l'=1}^{L-1} ) p( s_{d,t_{L}}^{[L]} \mid \mathcal{S}_{d,t_{L}-1}^{[L]}, \pi_{d,t_{L}}^{L,(m)})
\end{align}
The first term in this factorization is the likelihood. The second term is the state transition model, which describes how the current state \( \mathbf{x}_{d,t_{L}}^{[L]} \) evolves from the previous state \( \mathbf{x}_{d,t_{L}-1}^{[L]} \), and is influenced by the finer time scale states \( \left\{ \mathbf{x}^{[l']}_{d,1:T_{l'}} \right\}_{l'=1}^{L-1} \) and the indicator \( s_{d,t_{L}}^{[L]} \) corresponding to model $m$. The third term is the prior on the indicator \( s_{d,t_{L}}^{[L]} \), which depends on the previous indicators \( \mathcal{S}_{d,t_{L}-1}^{[L]} \) and the probabilities \( \pi_{d,t_{L}}^{L,(m)} \):
\begin{align}
    p( s_{d,t_{L}}^{[L]} \mid \mathcal{S}_{d,t_{L}-1}^{[L]}, \mathbf{x}_{d,t_{L}}^{[L]}, 
    \mathbf{y}_{d,t_{L}}^{[L]}, \pi_{d,t_{L}}^{L,(m)} ) \propto p( \mathbf{y}_{d,t_{L}}^{[L]} \mid \mathbf{x}_{d,t_{L}}^{[L]}) \nonumber \\
    p( \mathbf{x}_{d,t_{L}}^{[L]} \mid s_{d,t_{L}}^{[L]}, \mathbf{x}_{d,t_{L}-1}^{[L]}, 
    \left\{ \mathbf{x}^{[l']}_{d,1:T_{l'}} \right\}_{l'=1}^{L-1} ) \nonumber \\
    p( s_{d,t_{L}}^{[L]} \mid \mathcal{S}_{d,t_{L}-1}^{[L]}, \pi_{d,t_{L}}^{L,(m)})
\end{align}

\newpage
\subsection{Figures}

\newpage 
\begin{landscape}
\begin{longtable}{|c|c|c|c|c|c|c|c|c|c|c|c|c|c|c|c|}
\hline
\multirow{2}{*}{Individual} & \multicolumn{3}{c|}{Coarse Step 1} & \multicolumn{3}{c|}{Coarse Step 2} & \multicolumn{3}{c|}{Coarse Step 3} & \multicolumn{3}{c|}{Coarse Step 4} & \multicolumn{3}{c|}{Coarse Step 5} \\ \cline{2-16} 
 & Dim 1 & Dim 2 & Dim 3 & Dim 1 & Dim 2 & Dim 3 & Dim 1 & Dim 2 & Dim 3 & Dim 1 & Dim 2 & Dim 3 & Dim 1 & Dim 2 & Dim 3 \\ \hline
Individual 1 & 2.75e-16 & 2.75e-16 & 2.75e-16 & 0.123 & 0.0899 & 0.0931 & 0.0236 & 0.0208 & 0.0196 & 0.118 & 0.101 & 0.0752 & 0.0156 & 0.0141 & 0.0166 \\ \hline
Individual 2 & 2.75e-16 & 2.75e-16 & 2.75e-16 & 0.0806 & 0.105 & 0.0794 & 0.00999 & 0.0247 & 0.022 & 0.0677 & 0.0996 & 0.112 & 0.0111 & 0.0229 & 0.0198 \\ \hline
Individual 3 & 2.75e-16 & 2.75e-16 & 2.75e-16 & 0.0984 & 0.0952 & 0.143 & 0.0218 & 0.0272 & 0.0245 & 0.0813 & 0.107 & 0.0922 & 0.0176 & 0.0236 & 0.0182 \\ \hline
Individual 4 & 2.75e-16 & 2.75e-16 & 2.75e-16 & 0.0947 & 0.125 & 0.0917 & 0.0109 & 0.0272 & 0.0227 & 0.0804 & 0.112 & 0.109 & 0.0228 & 0.0139 & 0.0187 \\ \hline
Individual 5 & 2.75e-16 & 2.75e-16 & 2.75e-16 & 0.0602 & 0.103 & 0.0881 & 0.0146 & 0.0216 & 0.00997 & 0.0998 & 0.116 & 0.0928 & 0.0228 & 0.0282 & 0.0121 \\ \hline
Individual 6 & 2.75e-16 & 2.75e-16 & 2.75e-16 & 0.102 & 0.13 & 0.0725 & 0.0234 & 0.0201 & 0.0207 & 0.113 & 0.0871 & 0.0868 & 0.014 & 0.0141 & 0.0185 \\ \hline
\caption{Results for Coarse Steps 1 to 5 across dimensions for all individuals}
\label{tab:results_1_to_5}
\end{longtable}
% Table for Coarse Steps 6 to 10
\begin{longtable}{|c|c|c|c|c|c|c|c|c|c|c|c|c|c|c|c|}
\hline
\multirow{2}{*}{Individual} & \multicolumn{3}{c|}{Coarse Step 6} & \multicolumn{3}{c|}{Coarse Step 7} & \multicolumn{3}{c|}{Coarse Step 8} & \multicolumn{3}{c|}{Coarse Step 9} & \multicolumn{3}{c|}{Coarse Step 10} \\ \cline{2-16} 
 & Dim 1 & Dim 2 & Dim 3 & Dim 1 & Dim 2 & Dim 3 & Dim 1 & Dim 2 & Dim 3 & Dim 1 & Dim 2 & Dim 3 & Dim 1 & Dim 2 & Dim 3 \\ \hline
Individual 1 & 0.0955 & 0.0681 & 0.118 & 0.0214 & 0.0222 & 0.0235 & 0.121 & 0.0936 & 0.0816 & 0.0200 & 0.0168 & 0.0229 & 0.104 & 0.1029 & 0.0923 \\ \hline
Individual 2 & 0.1029 & 0.0823 & 0.0903 & 0.0232 & 0.0239 & 0.0208 & 0.0857 & 0.128 & 0.0756 & 0.0306 & 0.0197 & 0.0168 & 0.0899 & 0.1097 & 0.0926 \\ \hline
Individual 3 & 0.0838 & 0.0945 & 0.100 & 0.0195 & 0.0138 & 0.0185 & 0.1057 & 0.1014 & 0.116 & 0.0210 & 0.0093 & 0.0201 & 0.1026 & 0.1388 & 0.0711 \\ \hline
Individual 4 & 0.108 & 0.0637 & 0.0823 & 0.0203 & 0.0173 & 0.0270 & 0.100 & 0.0808 & 0.0845 & 0.0200 & 0.0093 & 0.0243 & 0.0983 & 0.111 & 0.0799 \\ \hline
Individual 5 & 0.0951 & 0.0807 & 0.124 & 0.0176 & 0.0216 & 0.0160 & 0.0834 & 0.1346 & 0.0862 & 0.0245 & 0.0208 & 0.0269 & 0.116 & 0.1136 & 0.0864 \\ \hline
Individual 6 & 0.111 & 0.0910 & 0.124 & 0.0303 & 0.0182 & 0.0203 & 0.0833 & 0.102 & 0.131 & 0.0239 & 0.0162 & 0.0206 & 0.0987 & 0.0844 & 0.112 \\ \hline
\caption{Results for Coarse Steps 6 to 10 across dimensions for all individuals}
\label{tab:results_6_to_10}
\end{longtable}

% Table for Coarse Steps 11 to 15
\begin{longtable}{|c|c|c|c|c|c|c|c|c|c|c|c|c|c|c|c|}
\hline
\multirow{2}{*}{Individual} & \multicolumn{3}{c|}{Coarse Step 11} & \multicolumn{3}{c|}{Coarse Step 12} & \multicolumn{3}{c|}{Coarse Step 13} & \multicolumn{3}{c|}{Coarse Step 14} & \multicolumn{3}{c|}{Coarse Step 15} \\ \cline{2-16} 
 & Dim 1 & Dim 2 & Dim 3 & Dim 1 & Dim 2 & Dim 3 & Dim 1 & Dim 2 & Dim 3 & Dim 1 & Dim 2 & Dim 3 & Dim 1 & Dim 2 & Dim 3 \\ \hline
Individual 1 & 0.0267 & 0.0124 & 0.0147 & 0.133 & 0.0927 & 0.0863 & 0.0175 & 0.0187 & 0.0291 & 0.123 & 0.0939 & 0.0751 & 0.0219 & 0.0227 & 0.0076 \\ \hline
Individual 2 & 0.0175 & 0.0233 & 0.0210 & 0.0635 & 0.128 & 0.101 & 0.0203 & 0.0141 & 0.0336 & 0.0693 & 0.0757 & 0.0965 & 0.0112 & 0.0090 & 0.0160 \\ \hline
Individual 3 & 0.0221 & 0.0138 & 0.0082 & 0.107 & 0.0996 & 0.0981 & 0.0271 & 0.0178 & 0.0132 & 0.104 & 0.106 & 0.106 & 0.0172 & 0.0258 & 0.0200 \\ \hline
Individual 4 & 0.0159 & 0.0085 & 0.0286 & 0.0963 & 0.0909 & 0.122 & 0.0085 & 0.0089 & 0.0211 & 0.0806 & 0.0752 & 0.0999 & 0.0150 & 0.0291 & 0.0160 \\ \hline
Individual 5 & 0.0174 & 0.0250 & 0.0084 & 0.104 & 0.100 & 0.0980 & 0.0089 & 0.0239 & 0.0432 & 0.0655 & 0.1157 & 0.122 & 0.0289 & 0.0119 & 0.0183 \\ \hline
Individual 6 & 0.0132 & 0.0191 & 0.0156 & 0.0821 & 0.129 & 0.0705 & 0.0184 & 0.0159 & 0.0188 & 0.0846 & 0.1176 & 0.109 & 0.0231 & 0.0129 & 0.0136 \\ \hline
\caption{Results for Coarse Steps 11 to 15 across dimensions for all individuals}
\label{tab:results_11_to_15}
\end{longtable}
\end{landscape}

\begin{landscape}
\begin{longtable}{|c|c|c|c|c|c|c|c|c|c|c|c|c|c|c|c|}
\hline
\multirow{2}{*}{Individual} & \multicolumn{3}{c|}{Coarse Step 16} & \multicolumn{3}{c|}{Coarse Step 17} & \multicolumn{3}{c|}{Coarse Step 18} & \multicolumn{3}{c|}{Coarse Step 19} & \multicolumn{3}{c|}{Coarse Step 20} \\ \cline{2-16} 
 & Dim 1 & Dim 2 & Dim 3 & Dim 1 & Dim 2 & Dim 3 & Dim 1 & Dim 2 & Dim 3 & Dim 1 & Dim 2 & Dim 3 & Dim 1 & Dim 2 & Dim 3 \\ \hline
Individual 1 & 0.119 & 0.102 & 0.0840 & 0.0236 & 0.0135 & 0.0339 & 0.0713 & 0.110 & 0.119 & 0.0250 & 0.0195 & 0.0203 & 0.117 & 0.102 & 0.0920 \\ \hline
Individual 2 & 0.0890 & 0.0863 & 0.113 & 0.0195 & 0.0300 & 0.0119 & 0.115 & 0.101 & 0.098 & 0.0202 & 0.0244 & 0.0159 & 0.0768 & 0.0928 & 0.116 \\ \hline
Individual 3 & 0.0944 & 0.0941 & 0.112 & 0.0071 & 0.0122 & 0.0268 & 0.0736 & 0.0851 & 0.118 & 0.0276 & 0.0167 & 0.0249 & 0.135 & 0.104 & 0.0660 \\ \hline
Individual 4 & 0.118 & 0.100 & 0.116 & 0.0176 & 0.0239 & 0.0195 & 0.0859 & 0.115 & 0.0752 & 0.0203 & 0.0118 & 0.0261 & 0.110 & 0.101 & 0.109 \\ \hline
Individual 5 & 0.116 & 0.0895 & 0.0930 & 0.0215 & 0.0333 & 0.00915 & 0.0830 & 0.130 & 0.0813 & 0.0182 & 0.0191 & 0.0162 & 0.0921 & 0.0862 & 0.110 \\ \hline
Individual 6 & 0.0878 & 0.105 & 0.0800 & 0.0330 & 0.0297 & 0.0103 & 0.0789 & 0.125 & 0.0971 & 0.0228 & 0.0084 & 0.0259 & 0.141 & 0.0607 & 0.0720 \\ \hline
\caption{Results for Coarse Steps 16 to 20 across dimensions for all individuals}
\label{tab:results_16_to_20}
\end{longtable}

% Table for Coarse Steps 21 to 25
\begin{longtable}{|c|c|c|c|c|c|c|c|c|c|c|c|c|c|c|c|}
\hline
\multirow{2}{*}{Individual} & \multicolumn{3}{c|}{Coarse Step 21} & \multicolumn{3}{c|}{Coarse Step 22} & \multicolumn{3}{c|}{Coarse Step 23} & \multicolumn{3}{c|}{Coarse Step 24} & \multicolumn{3}{c|}{Coarse Step 25} \\ \cline{2-16} 
 & Dim 1 & Dim 2 & Dim 3 & Dim 1 & Dim 2 & Dim 3 & Dim 1 & Dim 2 & Dim 3 & Dim 1 & Dim 2 & Dim 3 & Dim 1 & Dim 2 & Dim 3 \\ \hline
Individual 1 & 0.0189 & 0.0173 & 0.0189 & 0.0863 & 0.0856 & 0.0893 & 0.122 & 0.122 & 0.0847 & 0.0115 & 0.0253 & 0.0244 & 0.0831 & 0.117 & 0.0556 \\ \hline
Individual 2 & 0.0115 & 0.0245 & 0.0202 & 0.130 & 0.0687 & 0.0854 & 0.0226 & 0.0093 & 0.0170 & 0.0803 & 0.0643 & 0.129 & 0.0199 & 0.0175 & 0.00972 \\ \hline
Individual 3 & 0.0321 & 0.0199 & 0.00940 & 0.0996 & 0.0941 & 0.0779 & 0.0184 & 0.0166 & 0.0226 & 0.0848 & 0.0571 & 0.135 & 0.00906 & 0.0245 & 0.00895 \\ \hline
Individual 4 & 0.0818 & 0.120 & 0.0906 & 0.0204 & 0.0107 & 0.0124 & 0.0826 & 0.125 & 0.120 & 0.108 & 0.107 & 0.0973 & 0.0239 & 0.0158 & 0.0167 \\ \hline
Individual 5 & 0.0192 & 0.0150 & 0.0133 & 0.100 & 0.0790 & 0.109 & 0.0205 & 0.0137 & 0.0228 & 0.0992 & 0.111 & 0.101 & 0.0181 & 0.0174 & 0.0133 \\ \hline
Individual 6 & 0.0302 & 0.0235 & 0.00789 & 0.0974 & 0.0732 & 0.104 & 0.0298 & 0.0125 & 0.00930 & 0.0819 & 0.0979 & 0.0869 & 0.0188 & 0.0191 & 0.0181 \\ \hline
\caption{RMSE averaged across time scale $l=1$ for $t_{2} = 21,...,25$ across dimensions for all individuals}
\label{tab:results_21_to_25}
\end{longtable}

% Table for Coarse Steps 26 to 30
\begin{longtable}{|c|c|c|c|c|c|c|c|c|c|c|c|c|c|c|c|}
\hline
\multirow{2}{*}{Individual} & \multicolumn{3}{c|}{Coarse Step 26} & \multicolumn{3}{c|}{Coarse Step 27} & \multicolumn{3}{c|}{Coarse Step 28} & \multicolumn{3}{c|}{Coarse Step 29} & \multicolumn{3}{c|}{Coarse Step 30} \\ \cline{2-16} 
 & Dim 1 & Dim 2 & Dim 3 & Dim 1 & Dim 2 & Dim 3 & Dim 1 & Dim 2 & Dim 3 & Dim 1 & Dim 2 & Dim 3 & Dim 1 & Dim 2 & Dim 3 \\ \hline
Individual 1 & 0.0126 & 0.0167 & 0.0183 & 0.121 & 0.0905 & 0.0828 & 0.0243 & 0.0207 & 0.0220 & 0.0948 & 0.105 & 0.123 & 0.117 & 0.1018 & 0.0918 \\ \hline
Individual 2 & 0.0750 & 0.1117 & 0.0931 & 0.0110 & 0.0270 & 0.0198 & 0.0997 & 0.121 & 0.0902 & 0.0237 & 0.0207 & 0.0139 & 0.0968 & 0.0806 & 0.0802 \\ \hline
Individual 3 & 0.104 & 0.0793 & 0.100 & 0.0154 & 0.0272 & 0.0280 & 0.1149 & 0.0892 & 0.0803 & 0.0166 & 0.0176 & 0.0207 & 0.0879 & 0.0957 & 0.101 \\ \hline
Individual 4 & 0.107 & 0.0914 & 0.0880 & 0.0231 & 0.0147 & 0.0129 & 0.1120 & 0.0757 & 0.0916 & 0.0151 & 0.0159 & 0.0228 & 0.1098 & 0.0993 & 0.101 \\ \hline
Individual 5 & 0.0819 & 0.0846 & 0.101 & 0.0198 & 0.0136 & 0.0179 & 0.0879 & 0.0957 & 0.101 & 0.0303 & 0.0309 & 0.0265 & 0.1070 & 0.1025 & 0.0802 \\ \hline
Individual 6 & 0.1069 & 0.0925 & 0.0955 & 0.0128 & 0.0182 & 0.0243 & 0.0870 & 0.0892 & 0.131 & 0.0153 & 0.0241 & 0.0175 & 0.0975 & 0.0945 & 0.0788 \\ \hline
\caption{RMSE averaged across time scale $l=1$ for $t_{2} = 26,...,30$ across dimensions for all individuals}
\label{tab:results_26_to_30}
\end{longtable}
\end{landscape}
\begin{landscape}

\begin{longtable}{|c|c|c|c|c|c|c|c|c|c|c|c|c|c|c|c|}
\hline
\multirow{2}{*}{Individual} & \multicolumn{3}{c|}{Coarse Step 30} & \multicolumn{3}{c|}{Coarse Step 31} & \multicolumn{3}{c|}{Coarse Step 32} & \multicolumn{3}{c|}{Coarse Step 33} & \multicolumn{3}{c|}{Coarse Step 34} \\ \cline{2-16} 
 & Dim 1 & Dim 2 & Dim 3 & Dim 1 & Dim 2 & Dim 3 & Dim 1 & Dim 2 & Dim 3 & Dim 1 & Dim 2 & Dim 3 & Dim 1 & Dim 2 & Dim 3 \\ \hline
Individual 1 & 0.019 & 0.020 & 0.011 & 0.118 & 0.087 & 0.105 & 0.012 & 0.017 & 0.013 & 0.106 & 0.107 & 0.087 & 0.027 & 0.025 & 0.026 \\ \hline
Individual 2 & 0.134 & 0.073 & 0.106 & 0.018 & 0.015 & 0.021 & 0.099 & 0.111 & 0.087 & 0.019 & 0.020 & 0.021 & 0.083 & 0.108 & 0.126 \\ \hline
Individual 3 & 0.121 & 0.108 & 0.067 & 0.011 & 0.013 & 0.014 & 0.076 & 0.122 & 0.089 & 0.018 & 0.008 & 0.018 & 0.079 & 0.106 & 0.149 \\ \hline
Individual 4 & 0.073 & 0.052 & 0.073 & 0.102 & 0.130 & 0.112 & 0.009 & 0.018 & 0.017 & 0.124 & 0.086 & 0.112 & 0.016 & 0.028 & 0.020 \\ \hline
Individual 5 & 0.122 & 0.084 & 0.103 & 0.012 & 0.019 & 0.016 & 0.089 & 0.110 & 0.084 & 0.022 & 0.020 & 0.027 & 0.090 & 0.133 & 0.080 \\ \hline
Individual 6 & 0.108 & 0.084 & 0.095 & 0.017 & 0.015 & 0.013 & 0.118 & 0.119 & 0.091 & 0.017 & 0.021 & 0.012 & 0.089 & 0.125 & 0.093 \\ \hline
\caption{RMSE averaged across time scale $l=1$ for $t_{2} = 30,...,34$ across dimensions for all individuals}
\label{tab:results_30_to_34}
\end{longtable}
\begin{longtable}{|c|c|c|c|c|c|c|c|c|c|c|c|c|c|c|c|}
\hline
\multirow{2}{*}{Individual} & \multicolumn{3}{c|}{Coarse Step 35} & \multicolumn{3}{c|}{Coarse Step 36} & \multicolumn{3}{c|}{Coarse Step 37} & \multicolumn{3}{c|}{Coarse Step 38} & \multicolumn{3}{c|}{Coarse Step 39} \\ \cline{2-16} 
 & Dim 1 & Dim 2 & Dim 3 & Dim 1 & Dim 2 & Dim 3 & Dim 1 & Dim 2 & Dim 3 & Dim 1 & Dim 2 & Dim 3 & Dim 1 & Dim 2 & Dim 3 \\ \hline
Individual 1 & 0.119 & 0.094 & 0.101 & 0.022 & 0.014 & 0.016 & 0.129 & 0.095 & 0.093 & 0.026 & 0.020 & 0.016 & 0.105 & 0.093 & 0.115 \\ \hline
Individual 2 & 0.013 & 0.015 & 0.023 & 0.088 & 0.101 & 0.115 & 0.021 & 0.023 & 0.028 & 0.158 & 0.091 & 0.087 & 0.024 & 0.026 & 0.025 \\ \hline
Individual 3 & 0.018 & 0.010 & 0.013 & 0.071 & 0.100 & 0.105 & 0.011 & 0.012 & 0.013 & 0.072 & 0.108 & 0.123 & 0.023 & 0.031 & 0.025 \\ \hline
Individual 4 & 0.128 & 0.075 & 0.097 & 0.019 & 0.011 & 0.028 & 0.088 & 0.114 & 0.105 & 0.019 & 0.021 & 0.016 & 0.110 & 0.099 & 0.122 \\ \hline
Individual 5 & 0.034 & 0.012 & 0.020 & 0.106 & 0.093 & 0.122 & 0.028 & 0.023 & 0.011 & 0.100 & 0.110 & 0.129 & 0.027 & 0.022 & 0.011 \\ \hline
Individual 6 & 0.012 & 0.016 & 0.015 & 0.102 & 0.086 & 0.077 & 0.022 & 0.013 & 0.031 & 0.118 & 0.093 & 0.096 & 0.018 & 0.020 & 0.033 \\ \hline
\caption{RMSE averaged across time scale $l=1$ for $t_{2} = 35,...,39$ across dimensions for all individuals}
\label{tab:results_35_to_39}
\end{longtable}

\begin{longtable}{|c|c|c|c|c|c|c|c|c|c|c|c|c|c|c|c|}
\hline
\multirow{2}{*}{Individual} & \multicolumn{3}{c|}{Coarse Step 40} & \multicolumn{3}{c|}{Coarse Step 41} & \multicolumn{3}{c|}{Coarse Step 42} & \multicolumn{3}{c|}{Coarse Step 43} & \multicolumn{3}{c|}{Coarse Step 44} \\ \cline{2-16} 
 & Dim 1 & Dim 2 & Dim 3 & Dim 1 & Dim 2 & Dim 3 & Dim 1 & Dim 2 & Dim 3 & Dim 1 & Dim 2 & Dim 3 & Dim 1 & Dim 2 & Dim 3 \\ \hline
Individual 1 & 0.034 & 0.024 & 0.012 & 0.128 & 0.071 & 0.098 & 0.011 & 0.022 & 0.013 & 0.108 & 0.103 & 0.121 & 0.011 & 0.026 & 0.014 \\ \hline
Individual 2 & 0.106 & 0.084 & 0.102 & 0.016 & 0.033 & 0.030 & 0.092 & 0.129 & 0.104 & 0.015 & 0.032 & 0.020 & 0.090 & 0.098 & 0.089 \\ \hline
Individual 3 & 0.077 & 0.084 & 0.126 & 0.018 & 0.014 & 0.024 & 0.105 & 0.115 & 0.087 & 0.018 & 0.019 & 0.028 & 0.103 & 0.102 & 0.107 \\ \hline
Individual 4 & 0.020 & 0.030 & 0.030 & 0.106 & 0.107 & 0.085 & 0.011 & 0.012 & 0.015 & 0.088 & 0.079 & 0.095 & 0.020 & 0.028 & 0.021 \\ \hline
Individual 5 & 0.108 & 0.129 & 0.076 & 0.133 & 0.098 & 0.073 & 0.021 & 0.017 & 0.028 & 0.092 & 0.094 & 0.104 & 0.088 & 0.099 & 0.137 \\ \hline
Individual 6 & 0.120 & 0.099 & 0.113 & 0.019 & 0.018 & 0.024 & 0.122 & 0.101 & 0.089 & 0.019 & 0.030 & 0.024 & 0.112 & 0.094 & 0.085 \\ \hline
\caption{RMSE averaged across time scale $l=1$ for $t_{2} = 40,...,44$ across dimensions for all individuals}
\label{tab:results_40_to_44}
\end{longtable}

\end{landscape}

\begin{landscape}

\begin{longtable}{|c|c|c|c|c|c|c|c|c|c|c|c|c|c|c|c|}
\hline
\multirow{2}{*}{Individual} & \multicolumn{3}{c|}{Coarse Step 45} & \multicolumn{3}{c|}{Coarse Step 46} & \multicolumn{3}{c|}{Coarse Step 47} & \multicolumn{3}{c|}{Coarse Step 48} & \multicolumn{3}{c|}{Coarse Step 49} \\ \cline{2-16} 
 & Dim 1 & Dim 2 & Dim 3 & Dim 1 & Dim 2 & Dim 3 & Dim 1 & Dim 2 & Dim 3 & Dim 1 & Dim 2 & Dim 3 & Dim 1 & Dim 2 & Dim 3 \\ \hline
Individual 1 & 0.105 & 0.114 & 0.084 & 0.022 & 0.021 & 0.027 & 0.087 & 0.116 & 0.086 & 0.016 & 0.008 & 0.023 & 0.092 & 0.093 & 0.120 \\ \hline
Individual 2 & 0.013 & 0.037 & 0.033 & 0.098 & 0.090 & 0.121 & 0.110 & 0.095 & 0.109 & 0.013 & 0.016 & 0.015 & 0.082 & 0.113 & 0.095 \\ \hline
Individual 3 & 0.031 & 0.022 & 0.032 & 0.150 & 0.093 & 0.067 & 0.120 & 0.128 & 0.070 & 0.015 & 0.026 & 0.010 & 0.120 & 0.092 & 0.093 \\ \hline
Individual 4 & 0.090 & 0.113 & 0.093 & 0.015 & 0.022 & 0.011 & 0.110 & 0.084 & 0.104 & 0.023 & 0.029 & 0.011 & 0.087 & 0.118 & 0.085 \\ \hline
Individual 5 & 0.028 & 0.024 & 0.029 & 0.094 & 0.117 & 0.086 & 0.014 & 0.011 & 0.010 & 0.095 & 0.110 & 0.116 & 0.019 & 0.022 & 0.022 \\ \hline
Individual 6 & 0.027 & 0.024 & 0.011 & 0.118 & 0.083 & 0.095 & 0.023 & 0.021 & 0.030 & 0.091 & 0.108 & 0.116 & 0.014 & 0.013 & 0.021 \\ \hline
\caption{RMSE averaged across time scale $l=1$ for $t_{2} = 45,...,49$ across dimensions for all individuals}
\label{tab:results_45_to_49}
\end{longtable}

\begin{longtable}{|c|c|c|c|c|c|c|c|c|c|c|c|c|}
\hline
\multirow{2}{*}{Individual} & \multicolumn{3}{c|}{Coarse Step 50} & \multicolumn{3}{c|}{Coarse Step 51} & \multicolumn{3}{c|}{Coarse Step 52} & \multicolumn{3}{c|}{Coarse Step 53} \\ \cline{2-13} 
 & Dim 1 & Dim 2 & Dim 3 & Dim 1 & Dim 2 & Dim 3 & Dim 1 & Dim 2 & Dim 3 & Dim 1 & Dim 2 & Dim 3 \\ \hline
Individual 1 & 0.006 & 0.122 & 0.103 & 0.115 & 0.017 & 0.029 & 0.024 & 0.104 & 0.124 & 0.086 & 0.017 & 0.008 \\ \hline
Individual 2 & 0.011 & 0.093 & 0.140 & 0.090 & 0.016 & 0.019 & 0.008 & 0.098 & 0.080 & 0.064 & 0.016 & 0.030 \\ \hline
Individual 3 & 0.026 & 0.080 & 0.096 & 0.116 & 0.009 & 0.014 & 0.012 & 0.107 & 0.097 & 0.119 & 0.021 & 0.025 \\ \hline
Individual 4 & 0.018 & 0.121 & 0.072 & 0.093 & 0.020 & 0.014 & 0.026 & 0.074 & 0.153 & 0.073 & 0.014 & 0.015 \\ \hline
Individual 5 & 0.078 & 0.012 & 0.027 & 0.095 & 0.094 & 0.069 & 0.021 & 0.021 & 0.030 & 0.098 & 0.104 & 0.074 \\ \hline
Individual 6 & 0.092 & 0.014 & 0.010 & 0.033 & 0.116 & 0.086 & 0.087 & 0.029 & 0.024 & 0.019 & 0.089 & 0.117 \\ \hline
\caption{RMSE averaged across time scale $l=1$ for $t_{2} = 50,...,53$ across dimensions for all individuals}
\label{tab:results_50_to_53}
\end{longtable}

\begin{longtable}{|c|c|c|c|c|c|c|c|c|c|c|c|c|c|c|c|}
\hline
\multirow{2}{*}{Individual} & \multicolumn{3}{c|}{Coarse Step 54} & \multicolumn{3}{c|}{Coarse Step 55} & \multicolumn{3}{c|}{Coarse Step 56} & \multicolumn{3}{c|}{Coarse Step 57} & \multicolumn{3}{c|}{Coarse Step 58} \\ \cline{2-16} 
 & Dim 1 & Dim 2 & Dim 3 & Dim 1 & Dim 2 & Dim 3 & Dim 1 & Dim 2 & Dim 3 & Dim 1 & Dim 2 & Dim 3 & Dim 1 & Dim 2 & Dim 3 \\ \hline
Individual 1 & 0.017 & 0.008 & 0.019 & 0.083 & 0.095 & 0.110 & 0.027 & 0.010 & 0.027 & 0.095 & 0.112 & 0.120 & 0.027 & 0.012 & 0.021 \\ \hline
Individual 2 & 0.016 & 0.030 & 0.035 & 0.125 & 0.091 & 0.051 & 0.026 & 0.012 & 0.016 & 0.077 & 0.086 & 0.122 & 0.020 & 0.028 & 0.020 \\ \hline
Individual 3 & 0.021 & 0.025 & 0.029 & 0.077 & 0.086 & 0.085 & 0.017 & 0.030 & 0.027 & 0.091 & 0.108 & 0.075 & 0.019 & 0.022 & 0.020 \\ \hline
Individual 4 & 0.014 & 0.015 & 0.026 & 0.080 & 0.107 & 0.111 & 0.018 & 0.015 & 0.012 & 0.127 & 0.099 & 0.099 & 0.017 & 0.011 & 0.023 \\ \hline
Individual 5 & 0.098 & 0.104 & 0.074 & 0.008 & 0.018 & 0.024 & 0.104 & 0.126 & 0.102 & 0.024 & 0.031 & 0.017 & 0.068 & 0.088 & 0.126 \\ \hline
Individual 6 & 0.089 & 0.117 & 0.118 & 0.025 & 0.016 & 0.009 & 0.072 & 0.104 & 0.107 & 0.020 & 0.006 & 0.020 & 0.110 & 0.095 & 0.093 \\ \hline
\caption{RMSE averaged across time scale $l=1$ for $t_{2} = 54,...,58$ across dimensions for all individuals}
\label{tab:results_54_to_58}
\end{longtable}
\end{landscape}

\begin{landscape}
\begin{longtable}{|c|c|c|c|c|c|c|c|c|c|c|c|c|c|c|c|}
\hline
\multirow{2}{*}{Individual} & \multicolumn{3}{c|}{Coarse Step 59} & \multicolumn{3}{c|}{Coarse Step 60} & \multicolumn{3}{c|}{Coarse Step 61} & \multicolumn{3}{c|}{Coarse Step 62} & \multicolumn{3}{c|}{Coarse Step 63} \\ \cline{2-16} 
 & Dim 1 & Dim 2 & Dim 3 & Dim 1 & Dim 2 & Dim 3 & Dim 1 & Dim 2 & Dim 3 & Dim 1 & Dim 2 & Dim 3 & Dim 1 & Dim 2 & Dim 3 \\ \hline
Individual 1 & 0.125 & 0.075 & 0.080 & 0.008 & 0.019 & 0.028 & 0.071 & 0.110 & 0.124 & 0.019 & 0.021 & 0.022 & 0.080 & 0.112 & 0.125 \\ \hline
Individual 2 & 0.097 & 0.078 & 0.127 & 0.032 & 0.020 & 0.027 & 0.112 & 0.075 & 0.097 & 0.026 & 0.016 & 0.020 & 0.119 & 0.086 & 0.084 \\ \hline
Individual 3 & 0.092 & 0.080 & 0.120 & 0.015 & 0.010 & 0.015 & 0.112 & 0.076 & 0.124 & 0.020 & 0.013 & 0.009 & 0.164 & 0.069 & 0.095 \\ \hline
Individual 4 & 0.092 & 0.120 & 0.100 & 0.025 & 0.017 & 0.021 & 0.095 & 0.094 & 0.096 & 0.024 & 0.011 & 0.028 & 0.093 & 0.083 & 0.081 \\ \hline
Individual 5 & 0.024 & 0.022 & 0.017 & 0.100 & 0.101 & 0.083 & 0.033 & 0.024 & 0.029 & 0.107 & 0.091 & 0.097 & 0.016 & 0.016 & 0.015 \\ \hline
Individual 6 & 0.020 & 0.035 & 0.026 & 0.091 & 0.095 & 0.079 & 0.101 & 0.130 & 0.103 & 0.021 & 0.010 & 0.022 & 0.077 & 0.118 & 0.102 \\ \hline
\caption{RMSE averaged across time scale $l=1$ for $t_{2} = 59,...,63$ across dimensions for all individuals}
\label{tab:results_59_to_63}
\end{longtable}

\begin{longtable}{|c|c|c|c|c|c|c|c|c|c|c|c|c|c|c|c|}
\hline
\multirow{2}{*}{Individual} & \multicolumn{3}{c|}{Coarse Step 64} & \multicolumn{3}{c|}{Coarse Step 65} & \multicolumn{3}{c|}{Coarse Step 66} & \multicolumn{3}{c|}{Coarse Step 67} & \multicolumn{3}{c|}{Coarse Step 68} \\ \cline{2-16} 
 & Dim 1 & Dim 2 & Dim 3 & Dim 1 & Dim 2 & Dim 3 & Dim 1 & Dim 2 & Dim 3 & Dim 1 & Dim 2 & Dim 3 & Dim 1 & Dim 2 & Dim 3 \\ \hline
Individual 1 & 0.025 & 0.013 & 0.028 & 0.098 & 0.079 & 0.104 & 0.013 & 0.022 & 0.020 & 0.106 & 0.111 & 0.075 & 0.020 & 0.010 & 0.028 \\ \hline
Individual 2 & 0.022 & 0.030 & 0.022 & 0.097 & 0.113 & 0.070 & 0.024 & 0.020 & 0.012 & 0.088 & 0.089 & 0.137 & 0.031 & 0.016 & 0.019 \\ \hline
Individual 3 & 0.019 & 0.022 & 0.026 & 0.106 & 0.061 & 0.070 & 0.009 & 0.027 & 0.014 & 0.139 & 0.099 & 0.061 & 0.085 & 0.092 & 0.132 \\ \hline
Individual 4 & 0.018 & 0.027 & 0.017 & 0.084 & 0.131 & 0.091 & 0.023 & 0.008 & 0.013 & 0.103 & 0.102 & 0.112 & 0.010 & 0.010 & 0.020 \\ \hline
Individual 5 & 0.098 & 0.070 & 0.102 & 0.026 & 0.015 & 0.023 & 0.109 & 0.105 & 0.093 & 0.032 & 0.031 & 0.032 & 0.101 & 0.097 & 0.096 \\ \hline
Individual 6 & 0.017 & 0.011 & 0.010 & 0.125 & 0.107 & 0.078 & 0.017 & 0.011 & 0.032 & 0.090 & 0.090 & 0.106 & 0.018 & 0.020 & 0.023 \\ \hline
\caption{RMSE averaged across time scale $l=1$ for $t_{2} = 64,...,68$ across dimensions for all individuals}
\label{tab:results_64_to_68}
\end{longtable}
\begin{longtable}{|c|c|c|c|c|c|c|c|c|c|c|c|c|c|c|c|}
\hline
\multirow{2}{*}{Individual} & \multicolumn{3}{c|}{Coarse Step 69} & \multicolumn{3}{c|}{Coarse Step 70} & \multicolumn{3}{c|}{Coarse Step 71} & \multicolumn{3}{c|}{Coarse Step 72} & \multicolumn{3}{c|}{Coarse Step 73} \\ \cline{2-16} 
 & Dim 1 & Dim 2 & Dim 3 & Dim 1 & Dim 2 & Dim 3 & Dim 1 & Dim 2 & Dim 3 & Dim 1 & Dim 2 & Dim 3 & Dim 1 & Dim 2 & Dim 3 \\ \hline
Individual 1 & 0.111 & 0.112 & 0.112 & 0.122 & 0.106 & 0.094 & 0.021 & 0.016 & 0.018 & 0.079 & 0.081 & 0.071 & 0.025 & 0.011 & 0.018 \\ \hline
Individual 2 & 0.118 & 0.085 & 0.095 & 0.031 & 0.016 & 0.019 & 0.090 & 0.076 & 0.082 & 0.010 & 0.035 & 0.028 & 0.101 & 0.098 & 0.092 \\ \hline
Individual 3 & 0.008 & 0.022 & 0.025 & 0.076 & 0.124 & 0.094 & 0.022 & 0.017 & 0.019 & 0.113 & 0.096 & 0.107 & 0.020 & 0.021 & 0.016 \\ \hline
Individual 4 & 0.103 & 0.105 & 0.090 & 0.025 & 0.021 & 0.025 & 0.084 & 0.115 & 0.082 & 0.029 & 0.027 & 0.025 & 0.068 & 0.124 & 0.115 \\ \hline
Individual 5 & 0.023 & 0.019 & 0.036 & 0.069 & 0.133 & 0.080 & 0.023 & 0.023 & 0.016 & 0.093 & 0.084 & 0.102 & 0.023 & 0.024 & 0.013 \\ \hline
Individual 6 & 0.080 & 0.085 & 0.095 & 0.027 & 0.022 & 0.026 & 0.105 & 0.101 & 0.083 & 0.011 & 0.006 & 0.022 & 0.113 & 0.102 & 0.107 \\ \hline
\caption{RMSE averaged across time scale $l=1$ for $t_{2} = 69,...,73$ across dimensions for all individuals}
\label{tab:results_69_to_73}
\end{longtable}
\end{landscape}

\begin{landscape}
\begin{longtable}{|c|c|c|c|c|c|c|c|c|c|c|c|c|c|c|c|}
\hline
\multirow{2}{*}{Individual} & \multicolumn{3}{c|}{Coarse Step 74} & \multicolumn{3}{c|}{Coarse Step 75} & \multicolumn{3}{c|}{Coarse Step 76} & \multicolumn{3}{c|}{Coarse Step 77} & \multicolumn{3}{c|}{Coarse Step 78} \\ \cline{2-16} 
 & Dim 1 & Dim 2 & Dim 3 & Dim 1 & Dim 2 & Dim 3 & Dim 1 & Dim 2 & Dim 3 & Dim 1 & Dim 2 & Dim 3 & Dim 1 & Dim 2 & Dim 3 \\ \hline
Individual 1 & 0.119 & 0.093 & 0.085 & 0.018 & 0.024 & 0.010 & 0.101 & 0.101 & 0.082 & 0.012 & 0.014 & 0.022 & 0.103 & 0.114 & 0.099 \\ \hline
Individual 2 & 0.024 & 0.009 & 0.019 & 0.106 & 0.082 & 0.095 & 0.024 & 0.026 & 0.021 & 0.091 & 0.102 & 0.112 & 0.020 & 0.024 & 0.015 \\ \hline
Individual 3 & 0.089 & 0.087 & 0.086 & 0.097 & 0.111 & 0.108 & 0.036 & 0.021 & 0.008 & 0.101 & 0.109 & 0.084 & 0.015 & 0.015 & 0.021 \\ \hline
Individual 4 & 0.023 & 0.018 & 0.031 & 0.095 & 0.118 & 0.083 & 0.017 & 0.012 & 0.019 & 0.114 & 0.102 & 0.103 & 0.010 & 0.018 & 0.013 \\ \hline
Individual 5 & 0.105 & 0.087 & 0.097 & 0.016 & 0.016 & 0.014 & 0.102 & 0.126 & 0.096 & 0.020 & 0.011 & 0.023 & 0.095 & 0.108 & 0.081 \\ \hline
Individual 6 & 0.023 & 0.013 & 0.017 & 0.097 & 0.103 & 0.106 & 0.033 & 0.030 & 0.016 & 0.106 & 0.093 & 0.112 & 0.021 & 0.007 & 0.020 \\ \hline
\caption{RMSE averaged across time scale $l=1$ for $t_{2} = 74,...,78$ across dimensions for all individuals}
\label{tab:results_74_to_78}
\end{longtable}
\begin{longtable}{|c|c|c|c|c|c|c|c|c|c|c|c|c|c|c|c|}
\hline
\multirow{2}{*}{Individual} & \multicolumn{3}{c|}{Coarse Step 79} & \multicolumn{3}{c|}{Coarse Step 80} & \multicolumn{3}{c|}{Coarse Step 81} & \multicolumn{3}{c|}{Coarse Step 82} & \multicolumn{3}{c|}{Coarse Step 83} \\ \cline{2-16} 
 & Dim 1 & Dim 2 & Dim 3 & Dim 1 & Dim 2 & Dim 3 & Dim 1 & Dim 2 & Dim 3 & Dim 1 & Dim 2 & Dim 3 & Dim 1 & Dim 2 & Dim 3 \\ \hline
Individual 1 & 0.009 & 0.012 & 0.015 & 0.099 & 0.098 & 0.113 & 0.015 & 0.022 & 0.020 & 0.107 & 0.111 & 0.076 & 0.033 & 0.024 & 0.017 \\ \hline
Individual 2 & 0.100 & 0.132 & 0.081 & 0.013 & 0.011 & 0.025 & 0.124 & 0.127 & 0.078 & 0.022 & 0.010 & 0.015 & 0.099 & 0.075 & 0.117 \\ \hline
Individual 3 & 0.078 & 0.104 & 0.095 & 0.017 & 0.020 & 0.016 & 0.096 & 0.083 & 0.114 & 0.023 & 0.024 & 0.014 & 0.079 & 0.111 & 0.113 \\ \hline
Individual 4 & 0.125 & 0.113 & 0.082 & 0.090 & 0.115 & 0.129 & 0.032 & 0.023 & 0.016 & 0.102 & 0.102 & 0.105 & 0.028 & 0.017 & 0.006 \\ \hline
Individual 5 & 0.019 & 0.013 & 0.027 & 0.103 & 0.120 & 0.100 & 0.018 & 0.027 & 0.014 & 0.104 & 0.097 & 0.101 & 0.019 & 0.023 & 0.021 \\ \hline
Individual 6 & 0.108 & 0.087 & 0.097 & 0.016 & 0.012 & 0.029 & 0.098 & 0.129 & 0.075 & 0.024 & 0.009 & 0.020 & 0.063 & 0.065 & 0.072 \\ \hline
\caption{RMSE averaged across time scale $l=1$ for $t_{2} = 79,...,83$ across dimensions for all individuals}
\label{tab:results_79_to_83}
\end{longtable}
\begin{longtable}{|c|c|c|c|c|c|c|c|c|c|c|c|c|c|c|c|}
\hline
\multirow{2}{*}{Individual} & \multicolumn{3}{c|}{Coarse Step 84} & \multicolumn{3}{c|}{Coarse Step 85} & \multicolumn{3}{c|}{Coarse Step 86} & \multicolumn{3}{c|}{Coarse Step 87} & \multicolumn{3}{c|}{Coarse Step 88} \\ \cline{2-16} 
 & Dim 1 & Dim 2 & Dim 3 & Dim 1 & Dim 2 & Dim 3 & Dim 1 & Dim 2 & Dim 3 & Dim 1 & Dim 2 & Dim 3 & Dim 1 & Dim 2 & Dim 3 \\ \hline
Individual 1 & 0.106 & 0.102 & 0.085 & 0.024 & 0.021 & 0.014 & 0.098 & 0.124 & 0.096 & 0.018 & 0.018 & 0.019 & 0.101 & 0.122 & 0.075 \\ \hline
Individual 2 & 0.012 & 0.010 & 0.016 & 0.094 & 0.087 & 0.099 & 0.026 & 0.026 & 0.020 & 0.082 & 0.118 & 0.115 & 0.022 & 0.019 & 0.023 \\ \hline
Individual 3 & 0.019 & 0.023 & 0.016 & 0.130 & 0.105 & 0.075 & 0.011 & 0.018 & 0.015 & 0.095 & 0.099 & 0.110 & 0.024 & 0.030 & 0.018 \\ \hline
Individual 4 & 0.115 & 0.090 & 0.110 & 0.123 & 0.103 & 0.106 & 0.019 & 0.024 & 0.016 & 0.115 & 0.087 & 0.098 & 0.020 & 0.015 & 0.015 \\ \hline
Individual 5 & 0.093 & 0.131 & 0.096 & 0.012 & 0.020 & 0.021 & 0.117 & 0.086 & 0.095 & 0.128 & 0.108 & 0.080 & 0.009 & 0.018 & 0.013 \\ \hline
Individual 6 & 0.017 & 0.021 & 0.017 & 0.090 & 0.122 & 0.104 & 0.018 & 0.020 & 0.020 & 0.101 & 0.096 & 0.082 & 0.007 & 0.016 & 0.015 \\ \hline
\caption{RMSE averaged across time scale $l=1$ for $t_{2} = 84,...,88$ across dimensions for all individuals}
\label{tab:results_84_to_88}
\end{longtable}

\end{landscape}

\begin{landscape}
\begin{longtable}{|c|c|c|c|c|c|c|c|c|c|c|c|c|c|c|c|}
\hline
\multirow{2}{*}{Individual} & \multicolumn{3}{c|}{Coarse Step 89} & \multicolumn{3}{c|}{Coarse Step 90} & \multicolumn{3}{c|}{Coarse Step 91} & \multicolumn{3}{c|}{Coarse Step 92} & \multicolumn{3}{c|}{Coarse Step 93} \\ \cline{2-16} 
 & Dim 1 & Dim 2 & Dim 3 & Dim 1 & Dim 2 & Dim 3 & Dim 1 & Dim 2 & Dim 3 & Dim 1 & Dim 2 & Dim 3 & Dim 1 & Dim 2 & Dim 3 \\ \hline
Individual 1 & 0.026 & 0.013 & 0.017 & 0.063 & 0.129 & 0.085 & 0.016 & 0.017 & 0.010 & 0.114 & 0.122 & 0.084 & 0.150 & 0.091 & 0.097 \\ \hline
Individual 2 & 0.080 & 0.089 & 0.093 & 0.020 & 0.025 & 0.023 & 0.085 & 0.102 & 0.097 & 0.016 & 0.014 & 0.008 & 0.097 & 0.106 & 0.106 \\ \hline
Individual 3 & 0.093 & 0.067 & 0.131 & 0.007 & 0.023 & 0.010 & 0.056 & 0.074 & 0.096 & 0.024 & 0.009 & 0.022 & 0.104 & 0.074 & 0.108 \\ \hline
Individual 4 & 0.084 & 0.091 & 0.076 & 0.022 & 0.014 & 0.020 & 0.079 & 0.101 & 0.114 & 0.012 & 0.023 & 0.019 & 0.123 & 0.114 & 0.089 \\ \hline
Individual 5 & 0.090 & 0.135 & 0.093 & 0.140 & 0.097 & 0.092 & 0.024 & 0.016 & 0.017 & 0.110 & 0.090 & 0.091 & 0.028 & 0.021 & 0.014 \\ \hline
Individual 6 & 0.109 & 0.107 & 0.107 & 0.100 & 0.123 & 0.060 & 0.024 & 0.023 & 0.009 & 0.099 & 0.090 & 0.126 & 0.013 & 0.014 & 0.033 \\ \hline
\caption{RMSE averaged across time scale $l=1$ for $t_{2} = 89,...,93$ across dimensions for all individuals}
\label{tab:results_89_to_93}
\end{longtable}
\begin{longtable}{|c|c|c|c|c|c|c|c|c|c|c|c|c|c|c|c|}
\hline
\multirow{2}{*}{Individual} & \multicolumn{3}{c|}{Coarse Step 94} & \multicolumn{3}{c|}{Coarse Step 95} & \multicolumn{3}{c|}{Coarse Step 96} & \multicolumn{3}{c|}{Coarse Step 97} \\ \cline{2-16} 
 & Dim 1 & Dim 2 & Dim 3 & Dim 1 & Dim 2 & Dim 3 & Dim 1 & Dim 2 & Dim 3 & Dim 1 & Dim 2 & Dim 3 \\ \hline
Individual 1 & 0.025 & 0.014 & 0.017 & 0.084 & 0.101 & 0.115 & 0.021 & 0.018 & 0.014 & 0.092 & 0.084 & 0.083 \\ \hline
Individual 2 & 0.018 & 0.028 & 0.019 & 0.091 & 0.092 & 0.118 & 0.025 & 0.019 & 0.016 & 0.103 & 0.107 & 0.097 \\ \hline
Individual 3 & 0.018 & 0.016 & 0.017 & 0.080 & 0.099 & 0.100 & 0.020 & 0.025 & 0.017 & 0.060 & 0.092 & 0.079 \\ \hline
Individual 4 & 0.016 & 0.010 & 0.014 & 0.101 & 0.114 & 0.097 & 0.016 & 0.030 & 0.024 & 0.119 & 0.071 & 0.094 \\ \hline
Individual 5 & 0.105 & 0.113 & 0.084 & 0.023 & 0.013 & 0.013 & 0.116 & 0.101 & 0.096 & 0.026 & 0.026 & 0.027 \\ \hline
Individual 6 & 0.112 & 0.087 & 0.096 & 0.102 & 0.141 & 0.099 & 0.009 & 0.021 & 0.010 & 0.089 & 0.105 & 0.109 \\ \hline
\caption{RMSE averaged across time scale $l=1$ for $t_{2} = 94,...,97$ across dimensions for all individuals}
\label{tab:results_94_to_97}
\end{longtable}
\begin{longtable}{|c|c|c|c|c|c|c|c|c|c|}
\hline
\multirow{2}{*}{Individual} & \multicolumn{3}{c|}{Coarse Step 98} & \multicolumn{3}{c|}{Coarse Step 99} & \multicolumn{3}{c|}{Coarse Step 100} \\ \cline{2-10} 
 & Dim 1 & Dim 2 & Dim 3 & Dim 1 & Dim 2 & Dim 3 & Dim 1 & Dim 2 & Dim 3 \\ \hline
Individual 1 & 0.022 & 0.022 & 0.038 & 0.081 & 0.098 & 0.121 & 0.014 & 0.024 & 0.024 \\ \hline
Individual 2 & 0.014 & 0.031 & 0.021 & 0.112 & 0.102 & 0.095 & 0.010 & 0.021 & 0.008 \\ \hline
Individual 3 & 0.017 & 0.016 & 0.029 & 0.094 & 0.082 & 0.112 & 0.016 & 0.021 & 0.029 \\ \hline
Individual 4 & 0.020 & 0.019 & 0.013 & 0.116 & 0.077 & 0.094 & 0.011 & 0.013 & 0.015 \\ \hline
Individual 5 & 0.102 & 0.138 & 0.076 & 0.019 & 0.028 & 0.020 & 0.081 & 0.124 & 0.107 \\ \hline
Individual 6 & 0.023 & 0.020 & 0.014 & 0.143 & 0.092 & 0.079 & 0.009 & 0.009 & 0.022 \\ \hline
\caption{RMSE averaged across time scale $l=1$ for $t_{2} = 98,...,100$ across dimensions for all individuals}
\label{tab:results_98_to_100}
\end{longtable}

\end{landscape}

\end{document}